\documentclass{article}

\usepackage{microtype}
\usepackage{graphicx}
\usepackage{subfigure}
\usepackage{booktabs} 

\usepackage{booktabs}
\usepackage{multirow}
\usepackage[table]{xcolor}
\usepackage{tabularx}
\usepackage{nicematrix}
\usepackage{balance}

\usepackage{hyperref}

\usepackage[accepted]{icml2024}

\usepackage{amsmath}
\usepackage{amssymb}
\usepackage{mathtools}
\usepackage{amsthm}

\usepackage[capitalize,noabbrev]{cleveref}

\usepackage[capitalize]{cleveref}
\DeclareMathOperator*{\argmax}{arg\,max}

\crefname{section}{Sec.}{Secs.}
\Crefname{section}{Section}{Sections}
\Crefname{table}{Table}{Tables}
\crefname{table}{Tab.}{Tabs.}

\newcommand{\ieno}{\textit{i}.\textit{e}.}
\newcommand{\egno}{\textit{e}.\textit{g}.} 

\newcommand{\tcr}{\textcolor{black}}
\newcommand{\lsh}{\textcolor{black}}

\newcommand{\jx}{\textcolor{black}}

\theoremstyle{plain}

\theoremstyle{definition}

\theoremstyle{remark}

\usepackage[textsize=tiny]{todonotes}

\icmltitlerunning{StyDeSty: Min-Max Stylization and Destylization for Single Domain Generalization}

\begin{document}

\twocolumn[
\icmltitle{StyDeSty: Min-Max Stylization and Destylization\\for Single Domain Generalization}



\icmlsetsymbol{equal}{*}

\begin{icmlauthorlist}
\icmlauthor{Songhua Liu}{nus}
\icmlauthor{Xin Jin}{eias,nus}
\icmlauthor{Xingyi Yang}{nus}
\icmlauthor{Jingwen Ye}{nus}
\icmlauthor{Xinchao Wang}{nus}
\end{icmlauthorlist}

\icmlaffiliation{nus}{National University of Singapore, Singapore}
\icmlaffiliation{eias}{Eastern Institute of Technology, Ningbo, China}

\icmlcorrespondingauthor{Xinchao Wang}{xinchao@nus.edu.sg}

\icmlkeywords{Machine Learning, ICML}

\vskip 0.3in
]



\printAffiliationsAndNotice{}  

\begin{abstract}
  Single domain generalization (single DG) aims at learning a robust model generalizable to unseen domains from only one training domain, making it a highly ambitious and challenging task.
  State-of-the-art approaches have mostly relied on data augmentations, such as adversarial perturbation and style enhancement, to synthesize new data and thus increase robustness. 
  Nevertheless, they have largely overlooked the underlying coherence between the augmented domains, which in turn leads to inferior results in real-world scenarios.
  In this paper, we propose a simple yet effective scheme, termed as \emph{StyDeSty}, to explicitly account for the alignment of the source and pseudo domains in the process of data augmentation, enabling them to interact with each other in a self-consistent manner and further giving rise to a latent domain with strong generalization power.
  The heart of StyDeSty lies in the interaction between a \emph{stylization} module for generating novel stylized samples using the source domain, and a \emph{destylization} module for transferring stylized and source samples to a latent domain to learn content-invariant features.
  The stylization and destylization modules work adversarially and reinforce each other.
  During inference, the destylization module transforms the input sample with an arbitrary style shift to the latent domain, in which the downstream tasks are carried out. 
  Specifically, the location of the destylization layer within the backbone network is determined by a dedicated neural architecture search (NAS) strategy. 
  We evaluate StyDeSty on multiple benchmarks and demonstrate that it yields encouraging results, outperforming the state of the art by up to {13.44\%} on classification accuracy.  
  Codes are available \href{https://github.com/Huage001/StyDeSty}{here}. 
\end{abstract}

\section{Introduction}
\label{sec:intro}

\begin{figure}[t]
    \centering
    \includegraphics[width=\linewidth]{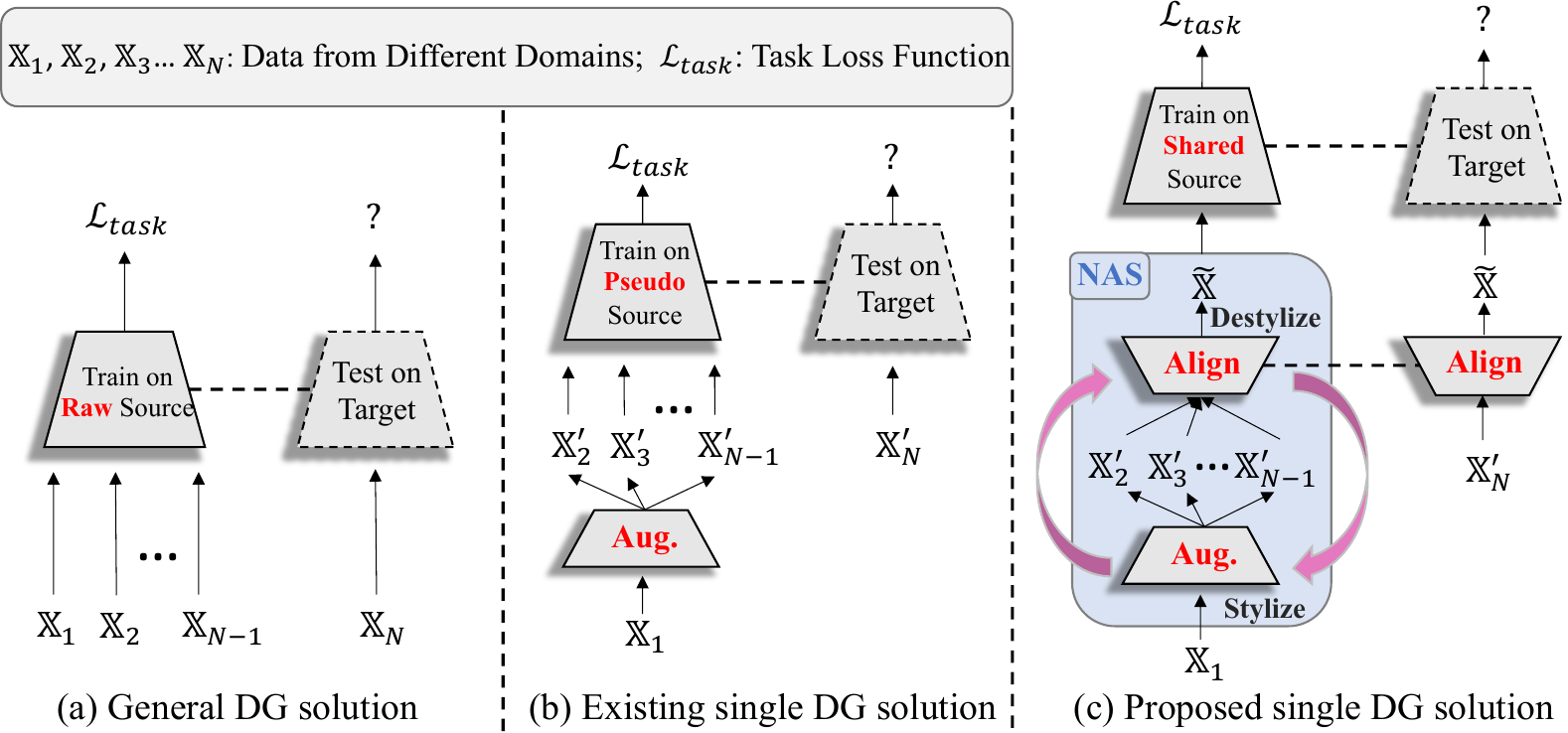}
    \vspace{-0.8cm}
    \caption{Workflows of general DG solution, 
    existing single DG solution, 
    and the proposed solution. 
    (a) General DG methods are trained on multiple source domains to learn domain-invariant representations for generalization. 
    (b) Existing single DG methods typically leverage data augmentation techniques to increase the domain diversity and then conduct training directly on the pseudo domains. 
    (c) The proposed single DG solution enables
    an explicit \emph{stylization} and \emph{destylization} mechanisms
    to learn a latent domain, where the downstream tasks are performed.
    The stylization and destylization work in an adversarial fashion, and the location of destylization
    is determined by a NAS algorithm.}
    \vspace{-0.5cm}
    \label{fig:motivation}
\end{figure}

Domain generalization (DG)~\cite{wang2021generalizing,sinha2017certifying,volpi2019addressing,volpi2018generalizing}
aims to tackle the 
distribution shift problem
between source and target domain,
and has recently demonstrated unprecedentedly
promising results.
The conventional setup of DG,
as shown in \cref{fig:motivation}(a),
includes  
multiple source domains $\mathbb{X}_1,\mathbb{X}_2,\cdots,\mathbb{X}_N$
during training and
a novel domain during testing.
As such, standard DG approaches have largely relied on
learning a domain-invariant representation
from the given domains,
so that the model can be successfully
generalized to other unseen domains. 
Nevertheless, 
access to data from multiple domains
is, in reality, often infeasible
due to data availability such as
privacy and budgeting issues.
This further calls for the 
\emph{single domain generalization}~(single DG)
that handles the generation-learning task
using only one source domain~\cite{qiao2020learning}.

Unfortunately, off-the-shelf solutions to multi-domain DG 
are not applicable to single DG,
since the former relies on 
domain identifiers as supervision signals for 
learning domain-invariant 
representation~\cite{balaji2018metareg,chattopadhyay2020learning,dou2019domain,li2019episodic},
which are however not available in the latter. 
To address the more challenging single DG task,
existing methods have resorted to 
data augmentation techniques to 
enhance the data diversity, 
including adversarial data
pertubation~\cite{qiao2020learning,volpi2018generalizing} and style enhancement~\cite{wang2021learning,zhou2020learning},
as shown in \cref{fig:motivation}(b).
Despite unprecedented advances,
existing endeavors have  
focused on the generation of the augmented domains,
yet largely overlooked
the interconnections between such pseudo-domains
and the source one,
leading to the incompetent generalization capability and 
further inferior results,
especially in the challenging in-the-wild scenarios.

In this paper, we introduce a novel single DG approach, 
termed as \emph{StyDeSty}, 
to explicitly explore and take advantage of
the underlying coherence between
the source and augmented domains,
in aim to learn a latent domain with strong
generalization capability. 
The core idea of StyDeSty lies
in that, samples from the pseudo-domains
should share the same underlying 
distribution in a ``hidden'' domain,
denoted as $\widetilde{\mathbb{X}}$,
with the source. 
This hidden domain $\widetilde{\mathbb{X}}$, intuitively,
resembles the \emph{content} in style-transfer tasks;
in other words, despite the diversified 
stylizations of  
$\mathbb{X'}_2,\mathbb{X'}_3,\cdots,\mathbb{X'}_{N-1}$,
their contents are all identical.
Such domain-invariant content is, therefore, reasonably expected to be generalized well to
unseen domains, which again can be treated
as unknown stylizations upon the same content.

We show in \cref{fig:motivation}(c)
the overall pipeline of the proposed StyDeSty.
It comprises three key components:
a \emph{stylization} module, 
a \emph{destylization} module, 
and a task head,
in which the former two are 
optimized in a min-max game to reinforce each other.
During training, the stylization module learns to 
generate novel stylized samples for the source domain,
while the destylization 
learns to unify the features before and after stylization
to an identical distribution in the latent domain,
and explicitly enforces 
content consistency between features of style-augmented samples and original ones.
During inference, testing samples,
treated as unseen-stylized ones,
are projected back to the learned latent domain,
where the downstream tasks, such as classifications, are performed through the task-specific head. 

\lsh{
Specifically, we first demonstrate the benefit of explicit destylization with an illustrating example. 
Then, we explore what is an appropriate objective to regulate the behavior of destylization, which provides insights for the adopted training algorithm.  
Finally, we reveal that the location of the key destylization layer is one crucial factor that affects the performance and devise a neural architecture search (NAS) strategy to automatically identify the \jx{optimal} location. 
In other words, in this paper, we study and give solutions to three questions: \emph{\jx{why}, how, and where to destyle in single DG?}
}
As demonstrated in our experiments, StyDeSty outperforms state-of-the-art models by $3.60\%$, $5.65\%$, and $13.44\%$ on Digits, CIFAR-10-C, and PACS respectively. 
Moreover, 
unlike previous DG techniques,
StyDeSty does not \lsh{require a specific label format like categorical data,} 
which makes it a versatile solution for not only classification but also regression, such as depth estimation. 

Our main contributions are thus summarized as follows:
\begin{itemize}
    \item A novel single DG approach \jx{is introduced}, termed as StyDeSty, 
    to explicitly account for the alignment of the source and pseudo domains, achieved
    through an adversarial stylization and destylization game,
    where the two players reinforce each other.
    \item \lsh{\jx{An} effective objective \jx{is further proposed as supervision} for the destylization module, \jx{along with} the corresponding training \jx{strategy} for the \jx{entire} framework.} 
    
    \item A NAS approach \jx{is devised} to identify the \jx{optimal} location for destylization,
    which \jx{coordinates well} \lsh{stylizations, destylizations, and downstream tasks}. 
    \item The proposed StyDeSty serves as a versatile solution for universal supervised learning problems, applicable to not only classification but also regression tasks. 
    Our method is evaluated extensively on multiple benchmarks and achieves superior performance. 
\end{itemize}

\section{Related Works}

\subsection{Alignment in General Domain Generalization}

The key for generalization lies in learning domain-invariant representation so that different domains share a common \tcr{feature/latent} space~\cite{ben2010theory}. 
Thus, many methods rely on aligning the source domains \tcr{by} minimizing cross-domain feature difference~\cite{motiian2017unified,RunpengCVPR2023,jin2020feature,mahajan2021domain,du2020learning,XingyiECCV22,XingyiNeurIPS22,JingwenPNC2023,JingwenCVPR24Ungeneralizable,liu2022dataset,muandet2013domain}, \egno, based on distance metrics of maximum mean discrepancy~(MMD)~\cite{long2015learning,yan2017mind}, correlation distance~\cite{sun2016deep,zhuo2017deep}, and second-order moment~\cite{peng2019moment,jin2020feature}. The other branch that achieves a similar goal tends to leverage the adversarial training strategy for generalization~\cite{ganin2015unsupervised,li2018deep,li2018domain,shao2019multi,zhao2020domain,albuquerque2019generalizing}. Although these methods could handle the general DG problem, they are still inapplicable to single DG where only one source domain is available since both of the distance minimization and adversarial training based DG methods need multiple source domains for optimization. 

On the other routine, there are also works designing normalization mechanisms for feature alignment so that \tcr{features from different domains} share the same statistics. For example, \cite{seo2020learning}~\tcr{propose to learn the domain-specific batch normalization layer for each domain independently. \cite{fan2021adversarially} present an adversarial adaptive normalization where both the standardization and rescaling statistics are learned via neural networks instead of data-wise calculation.} \cite{jin2021style} design a restoration module that supplements the lost discriminative features due to the normalization operation. Differently, the feature destylization design in our paper is achieved by the adaptive instance normalization (AdaIN)~\cite{huang2017arbitrary} which aims to transfer all the augmented and stylized features to the same distribution, so as to reach ``alignment''. Moreover, destylization also enforces constraints to preserve content consistency before and after stylization, which further encourages the learning of style-/domain-invariant features. 

We notice that \cite{DBLP:conf/aaai/Yang0023} also adopt the AdaIN module for style alignment. 
Nevertheless, the augmentation techniques are limited to low-level transformations such as color jittering and Gaussian noise, resulting in relatively limited data diversity. 
Furthermore, their augmentation and alignment modules are trained independently. 
In contrast, the two modules learn mutually to reinforce each other in our method. 
We find that such an adversarial fashion impacts performance positively. 

\subsection{Single Domain Generalization}

The problem of learning to generalize with only one available source domain can be tackled from different perspectives, such as projecting superficial statistics out~\cite{wang2019learning1}, penalizing local predictive power~\cite{wang2019learning2}, solving Jigsaw puzzles~\cite{carlucci2019domain}, clustering pseudo training domains~\cite{matsuura2020domain}, self challenging~\cite{huang2020self}, meta architectures~\cite{wan2022meta}, attention consistency~\cite{cugu2022attention}, and debiasing and regularization~\cite{qu2023modality}. 

However, recent data augmentation-based methods have achieved dominant performance for the single DG task by enriching the diversity of domain data. Representatively, \cite{volpi2018generalizing} propose to apply adversarial perturbations into the source samples to augment domain data. \cite{zhao2020maximum} further add regularization to adversarial perturbations via maximum entropy. 
\cite{li2021progressive} and \cite{kang2022style} propose to generate novel styles progressively and constantly. 
\cite{cheng2023adversarial} propose Beyasian augmentation. 
\cite{lv2022causality} and \cite{chen2023meta} augment data by leveraging causal knowledge. 
\cite{qiao2020learning} make the learning with adversarial data augmentation become learnable and optimize it in a bi-level meta-learning framework. 
Similar ideas are also explored in \cite{fu2023styleadv}, \cite{chen2022maxstyle}, and \cite{zhong2022adversarial}. 
Beyond adversarial perturbations, many recent data augmentation-based single DG methods tend to use style transfer techniques for domain enlargement, \egno, \cite{zhou2020learning} train an image-to-image generator for each source domain to synthesize novel domains, \cite{zhou2021domain} mix up the styles of source images to generate new domain data, \cite{wang2021learning} propose a style complement module to increase the diversity of domain data, \cite{choi2023progressive} propose progressive random convolution for style augmentation, and \cite{zhao2022style} utilize farthest point sampling to select style vectors for style augmentation. 
However, these methods over-explore the domain augmentation (\ieno, stylization) but ignore the important effect of the \textbf{explicit feature alignment} and \textbf{the underlying coherence among augmented domains} w.r.t model generalization, which makes the trained model hard to be generalized to test domains with unseen styles in inference. This drawback motivates us to further consider unifying/aligning the distribution of augmented domains by destylization and thus increase model robustness against style variance. How to balance such two designs of stylization and destylization is also the focus of this paper.

\section{Methodology}

\begin{figure}[t]
    \centering
    \includegraphics[width=\linewidth]{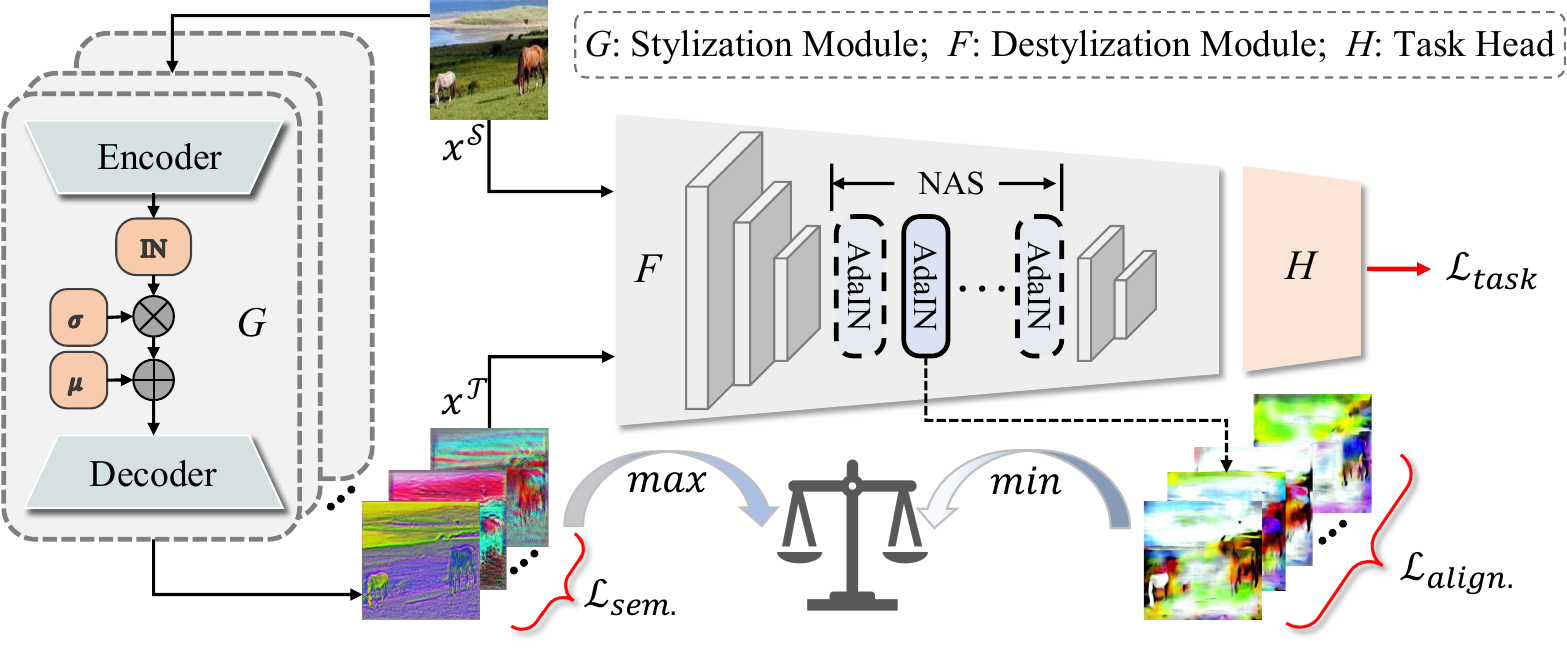}
    \vspace{-0.9cm}
    \caption{StyDeSty framework consists of a stylization module $G$, a destyliation module $F$, and a task head $H$, where a NAS algorithm is involved to search an optimal position of the AdaIN layer for destylization. Black and red arrows denote forward pass and loss computation and IN represents instance normalization.}
    \vspace{-0.6cm}
    \label{fig:pipeline}
\end{figure}

The heart of the StyDeSty framework is at the interaction of its three components: a stylization module $G$, a destylization module $F$, and a task \tcr{head} $H$. 
An overview of our method is illustrated in \cref{fig:pipeline}. 
We start the introduction of the key stylization and destylization modules first and then elaborate on the training objective. 
Finally, a NAS-guided training algorithm is proposed, which \tcr{coordinates the interaction between stylization and destylization, and regulates the whole StyDeSty pipeline.} 

\subsection{Stylization and Destylization}
\label{sec:stydesty}

\begin{figure*}[t]
    \centering
    \includegraphics[width=\linewidth]{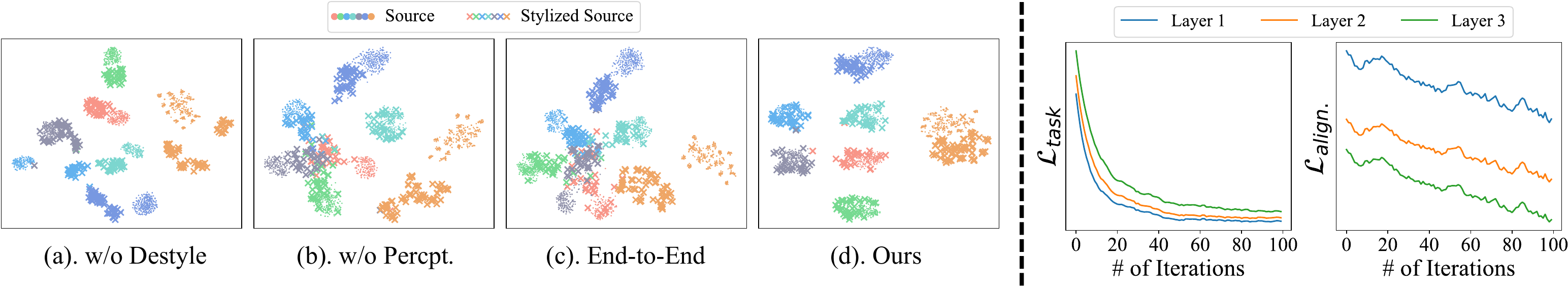}
    \vspace{-0.9cm}
    \caption{Left: TSNE visualizations of stylized and original source samples for by different fashions of destylization. Right: Visualizations of task loss and alignment loss with destylization at different locations in a deep network.}
    \vspace{-0.5cm}
    \label{fig:pre_exp}
\end{figure*}

\textbf{Stylization Module.} 
Similar to style transfer~\cite{DBLP:conf/cvpr/GatysEB16,huang2017arbitrary,DBLP:conf/iccv/LiuLHLWLSLD21,DBLP:conf/eccv/LiuYRW22,DBLP:conf/mm/LiuWLS20,DBLP:journals/corr/abs-2304-09728,DBLP:conf/cvpr/TangLLHLHW23}, the stylization module $G$ aims to generate various stylized versions given a source image $x^\mathcal{S}$ from a source domain $\mathcal{S}$. 
In this paper, similar to the style complement module in~\cite{wang2021learning}, $G$ consists of $B$ blocks, with a single convolution layer $enc_j$, an instance normalization layer, an affine transformation layer parameterized by $\mu_j$ and $\sigma_j$, and a symmetric deconvolution layer $dec_j$ for the $j$-th block, $1\leq j\leq B$. 
Given an input RGB image $x^\mathcal{S}$, the $j$-th block firstly projects it to a $c$-dimension feature space with $enc_j$ to derive $f_j\in \mathbb{R}^{h\times w\times c}$. 
Then, the instance normalization layer normalizes $f_j$ with the channel-independent mean $\mu^f_j$ and standard deviation $\sigma^f_j$, followed by the affine transformation layer to get $\hat{f_j}$. 
At last, $dec_j$ projects $\hat{f_j}$ back to the RGB space, and the result is denoted as $\hat{x_j}$. 
Formally, this process can be written as:
\begin{equation}
    f_j={\rm enc}_j(x^\mathcal{S}),\hspace{1mm} \hat{f_j}=\sigma_j\times\frac{f_j-\mu^f_j}{\sigma^f_j}+\mu_j,\hspace{1mm} \hat{x_j}={\rm dec}_j(\hat{f_j}). 
\end{equation}
Notably, the affine parameters $\mu_j$ and $\sigma_j$ for some blocks have shape $h\times w\times c$ while the other ones are with shape $1\times 1\times c$, to mimic local and global distortions respectively. 
The final augmented result $x^\mathcal{T}$ is given as a weighted sum of the outputs of all the $B$ blocks, with the weight vector $w\in\mathbb{R}^B$ drawn from a standard normal distribution in each training iteration:
\begin{equation}
    w\sim\mathcal{N}(0,1),\hspace{3mm} x^\mathcal{T}={\rm sigmoid}(\frac{1}{\sum_{j=1}^{B}w_j}w_j\hat{x_j}), \label{eq:2}
\end{equation}
where the ${\rm sigmoid}$ is applied to scale the augmented images.

\noindent\textbf{Destylization Module.}
The destylization module $F$ is typically composed of the first several blocks of a backbone network and a destylization layer. 
For example, for the classification problem, if \textit{ResNet-18}~\cite{he2016deep} with $4$ main blocks is selected as the backbone network, we can take the former $2$ blocks to build the destylization module and the remaining parts would serve as the task head $H$. 
The core of the destylization module is the final destylization layer accounting for \tcr{underlying coherence} and explicitly distribution aligning. In this paper, we adopt adaptive instance normalization (AdaIN)~\cite{huang2017arbitrary}, a simple but effective mechanism in arbitrary style transfer, \tcr{for instantiation}. 
The insight is that channel-wise statistic information like mean and variance in a deep network can largely represent the style of an image. 
In this sense, the alignment of these statistics can be viewed as transferring all the images to the same style/latent, which is known as \emph{destylization} in this paper. 
The unified style is encoded by the affine parameters of the AdaIN layer and the formulation is similar to that of the stylization module:
\begin{equation}
    {\rm AdaIN}(f,\mu,\sigma)=\sigma\times\frac{f-\mu^f}{\sigma^f}+\mu,
\end{equation}
where $\mu$ and $\sigma$ are learnable affine parameters while $\mu^f$ and $\sigma^f$ are the channel-wise mean and standard deviation of a feature map $f$ produced by the layer before AdaIN. 

\noindent\textbf{\jx{Discussion: \emph{Why to Destyle?}}} 
\lsh{
We provide an illustrating example in \cref{fig:pre_exp}(left) to demonstrate \jx{how our destylization module works}. 
We conduct training on the photo domain of PACS dataset~\cite{li2017deeper} and visualize the learned representation of each sample using TSNE~\cite{van2008visualizing}, where \jx{the} different classes are denoted by different colors, and \jx{the} original source samples and stylized ones are denoted by different markers. \jx{As shown in the plot (a), although the previous methods like L2D~\cite{wang2021learning} have used style augmentation to increase diversity, they still suffer from the domain shift problem given \emph{unseen} styles in inference without a kind of explicit alignment/destylization. In contrast, with an explicit destylization operation, our method largely alleviates such a problem (see plot (d)), which means better robustness to style shifts. More results can be found in the supplementary.}}

\subsection{Objective Functions}
\label{sec:objective}

\lsh{
\jx{The overall loss function is inspired by (1) the recent works on adversarial data augmentation}~\cite{sinha2017certifying,qiao2020learning,zhao2020maximum}, and (2) \jx{the classic theory of domain adaptation~\cite{ben2010theory}} \jx{--} the test error is largely dominated by \jx{the source training risk and the discrepancy between the source and target domain:}
}
\begin{equation}
\begin{aligned}
    \min\sup_{\mathcal{T}=G(\mathcal{S}):W(\mathcal{S},\mathcal{T})\leq\rho}\mathbb{E}_{\mathcal{T}}[&\mathcal{L}_{task}(H(F(x^\mathcal{T})),y^\mathcal{T})\\+\alpha&\mathcal{L}_{align.}(F(x^\mathcal{S}),F(x^\mathcal{T}))],\label{eq:ours}
\end{aligned}
\end{equation}
where $\mathcal{T}$ denotes synthetic target domains by data augmentation, $x^\mathcal{T}\sim\mathcal{T}$ is an instance sampled from $\mathcal{T}$ and augmented from the source sample $x^\mathcal{S}\sim\mathcal{S}$, $G$ and $F$ are instantiated as the stylization and destylization modules respectively in this paper, $\mathcal{L}_{align.}$ is a metric of alignment between two features, $\alpha$ is a hyperparameter controlling the weight of this constraint, and $\rho$ is another hyperparameter denoting the maximal strength of data augmentation. 
Since it is intractable for deep networks to solve the constrained optimization problem in \cref{eq:ours}, we alternatively consider the following objective by Lagrangian relaxation:
\begin{equation}
\begin{aligned}
    \min\sup_{\mathcal{T}=G(\mathcal{S})}\{\mathbb{E}_{\mathcal{T}}[&\mathcal{L}_{task}(H(F(x^\mathcal{T})),y^\mathcal{T})\\+\alpha\mathcal{L}_{align.}&(F(x^\mathcal{S}),F(x^\mathcal{T}))]-\beta W(\mathcal{S},\mathcal{T})\},\label{eq:relax}
\end{aligned}
\end{equation}
where $\beta\geq0$ is a penalty factor with an intuitive meaning similar to $\rho$. 
\cref{eq:relax} offers insights into the loss functions of each component which will be illustrated below. 

\noindent\textbf{Task head.} 
The task head $H$ in the StyDeSty framework aims to discriminate information related to the task from the unified distribution/latent by the destylization module $F$. 
Through \cref{eq:relax}, we can find that the feature alignment metric $\mathcal{L}_{align.}$ and the W-distance term are not related to $H$. 
Therefore, $H$ is trained with only the task-specific loss $\mathcal{L}_{task}(H(F(x^\mathcal{T})),y^\mathcal{T})$. 
Notably, StyDeSty is a versatile framework applicable for different tasks with different forms of $\mathcal{L}_{task}$. 
For instance, in classification problems, cross-entropy loss is a typical option, while in regression problems, we can use $L1$ or $L2$ loss as $\mathcal{L}_{task}$. 

\noindent\textbf{Destylization Module: \emph{How to Destyle}.} 
As suggested by \cref{eq:relax}, there are two components for the objective of the destylization module $F$: task-specific loss $\mathcal{L}_{task}$ and feature alignment metric $\mathcal{L}_{align.}$. 
The configuration of $\mathcal{L}_{task}$ is the same as that in the task head. 
As for the feature alignment term, one straight-forward idea is to measure the Euclidean distance or $L2$ distance between the two feature maps $F(x^\mathcal{S})$ and $F(x^\mathcal{T})$. 
However, \lsh{as shown in the plot (b) of \cref{fig:pre_exp}}, we find that this configuration often leads to inferior results. 
One major problem is that not all positions and channels are worth aligning equally for the current task. 
To increase the awareness of key features for this distance metric, we further adopt the task head as a perceptual network and measure the feature distance at the last hidden layer of $H$, denoted as $H_{-1}$. 
In this way, we have the following loss function for the destylization module:
\begin{equation}
    \begin{aligned}
        \mathcal{L}_{align.}&(F(x^\mathcal{S}),F(x^\mathcal{T}))=\Vert F(x^\mathcal{S})-F(x^\mathcal{T})\Vert^2\\+\lambda&\Vert H_{-1}(F(x^\mathcal{S}))-H_{-1}(F(x^\mathcal{T}))\Vert^2,\\
        \mathcal{L}_{F}=&\frac{1}{n}\sum_{i=1}^n\{\mathcal{L}_{task}(H(F(x_i^\mathcal{T})),y_i^\mathcal{T})\\&+\alpha\mathcal{L}_{align.}(F(x_i^\mathcal{S}),F(x_i^\mathcal{T}))\},\label{eq:loss_F}
    \end{aligned}
\end{equation}
where $\lambda$ is a hyper-parameter balancing the weight of the perceptual term and $n$ is the size of a mini-batch. 

It is worth noting that the task head $H$ serves as a metric function here and its parameters should not be updated according to the gradient of $\mathcal{L}_{align.}$. 
\lsh{
Therefore, we do not train modules $G$ and $H$ \jx{simultaneously} but update \jx{them} alternately. 
Otherwise, the task head would also help alignment, which weakens \jx{the alignment ability of the destylization module}, as demonstrated in the plot (c) of \cref{fig:pre_exp}. 
}

\noindent\textbf{Stylization Module.} 
The stylization module $G$ behaves adversarially against $F$ and $H$. \tcr{Moreover, it is not allowed to destroy the semantics of the original images to generate meaningless stylized results. 
Therefore, we introduce a semantic perceptron $M$ to enforce the semantic consistency constraint $\mathcal{L}_{sem.}$ on $G$. Besides,} the time complexity of solving the W distance as indicated by \cref{eq:relax} for a batch of data is considerable for an iterative algorithm. 
We then alternatively consider the dual form of the W distance, which is equivalent to the maximum mean discrepancy (MMD) with a Lipschitz continuous kernel function $k$ under some mild conditions~\cite{edwards2011kantorovich}. 
In this sense, the loss function for $G$ can be written as:
\begin{equation}
\begin{aligned}
    \mathcal{L}_{sem.}=\Vert\frac{1}{n}\sum_{i=1}^nk(M(x^\mathcal{S}_i))&-\frac{1}{n}\sum_{i=1}^nk(M(x^\mathcal{T}_i))\Vert^2,\\
    \mathcal{L}_{G}=-\mathcal{L}_{F}&+\beta\mathcal{L}_{sem.}.\label{eq:loss_G}
\end{aligned}
\end{equation}
\subsection{Training with NAS: \emph{Where to Destyle}}
\label{sec:nas}
With the different objective functions for each module, the training process can be organized as a three-stage algorithm, to train $F$, $H$, and $G$ alternately in each iteration. 
Nevertheless, we have to answer an important question before the formal training: how to select an appropriate position in a backbone network to insert the destylization layer and split the network into $F$ and $H$? 
\lsh{
Empirically, we observe that there is a trade-off between the objectives of the task ahead and the alignment. 
As demonstrated in \cref{fig:pre_exp}(right), \jx{the deeper destylization} could benefit the alignment of the source and stylized samples but \jx{make the task head training difficult}, since \jx{it is more convenient to enforce the features with high-level semantics to be aligned by discarding discriminative information.} 
}
In this paper, instead of selecting the position heuristically, we devise a neural architecture search (NAS) strategy to address this problem. 
Assume that there is a backbone network $P$ with $L$ positions that are potentially suitable for inserting the AdaIN layer which splits $P$ into $L+1$ blocks with the $l$-th one denoted as $p_l$. 
In the NAS stage, we insert AdaIN layers, denoted as ${\rm AdaIN}_l$ with $1\leq l\leq L$, to all the $L$ positions and optimize a vector $\pi\in\mathbb{R}^L$, where $\pi_l$ indicates the logit value that the $l$-th AdaIN layer is enabled. 
We denote the output of the $l$-th block as $x_l$, and $x_l$ for $1\leq l\leq L$ is given by:
\begin{equation}
\begin{aligned}
    \hat{\pi}={\rm GumbelSoftmax}&(\pi),
    \hspace{3mm}
    \hat{x_l}=p_l(x_{l-1}),\\
    x_l=\hat{\pi_l}{\rm AdaIN}_l&(\hat{x_l})+(1-\hat{\pi_l})\hat{x_l}.
\end{aligned}\label{eq:nas}
\end{equation}
Here, the Gumbel-Softmax function~\cite{jang2016categorical} is applied on $\pi$, which would produce a one-hot vector $\hat{\pi}$ indicating the selected AdaIN layer in this iteration. 

\begin{algorithm}[!t]
	\caption{Training of StyDeSty}
    \textbf{Required}: A source domain $\mathcal{S}$; A randomly-initialized stylization module $G$; A randomly initialized backbone $P$ with $L$ candidate AdaIN layers; A zero initialized vector $\pi$ for selecting the enabled AdaIN. 
    \begin{algorithmic}[1]
        \REPEAT
            \STATE Train $P$ and $\pi$ by \cref{eq:loss_nas_P} for $T_P$ times;
            \STATE Train $G$ by \cref{eq:loss_nas_G} for $T_G$ times;\\
        \UNTIL{$\argmax\pi$ does not change}\hfill\algorithmiccomment{NAS Stage}
        \STATE Select the AdaIN layer $l$ with maximal $\pi_l$ and split $P$ into $F$ and $H$;
        \REPEAT
            \STATE Train $F$ by \cref{eq:loss_F} for $T_F$ times;
            \STATE Train $H$ by $\mathcal{L}_{task}$ for $T_H$ times;
            \STATE Train $G$ by \cref{eq:loss_G} for $T_G$ times;
        \UNTIL{convergence}\hfill\algorithmiccomment{Formal training stage}
	\end{algorithmic}
	\label{alg:train}
\end{algorithm}

For optimization, all the parameters of $P$ including $\pi$ are updated together in the NAS time since the split position for the destylization module and the task head is unknown. 
$P$ and $G$ are still trained in a min-max game and their loss functions are formulated as:
\begin{equation}
\begin{aligned}
    \mathcal{L}_{P}=\frac{1}{n}\sum_{i=1}^n\{&\mathcal{L}_{task}(P(x_i^\mathcal{T}),y_i^\mathcal{T})\\+\alpha\sum_{l=1}^L\hat{\pi_l}\Vert {\rm AdaIN}_l&(\hat{x^\mathcal{S}_{i,l}})-{\rm AdaIN}_l(\hat{x^\mathcal{T}_{i,l}})\Vert^2\}\label{eq:loss_nas_P}, 
\end{aligned}
\end{equation}
\begin{equation}
    \mathcal{L}_{G}=-\mathcal{L}_{P}+\beta\mathcal{L}_{sem.}.\label{eq:loss_nas_G}
\end{equation}
Note that here we do not incorporate the perceptual term in \cref{eq:loss_F} which requires passing through features after each ${\rm AdaIN}_l$ till the last layer and increases the computational burden significantly. 
The NAS procedure will repeat until the selected index of AdaIN layer does not change in further iterations. 
The overall training is summarized as \cref{alg:train}. 
The memory complexity and time complexity per iteration are consistent with those augmentation-based DG methods like L2D~\cite{wang2021learning}, while the overall time complexity is related to $T_P$, $T_G$, $T_F$, and $T_H$. 

\section{Experiments}

\subsection{Datasets}
To demonstrate the effectiveness and versatility of the proposed StyDeSty for single DG, we conduct extensive evaluations on three classification benchmarks: \textit{Digits}, \textit{CIFAR-10-C}, and \textit{PACS}, and one regression problem: monocular depth estimation on the \textit{KITTI} and \textit{vKITTI} dataset. 
More comparisons can be found in the appendix. 

\noindent\textbf{Digits.} 
\textit{Digits} consists of $5$ digit recognition datasets including MNIST~\cite{lecun1998gradient}, SVHN~\cite{netzer2011reading}, MNIST-M~\cite{ganin2015unsupervised}, SYN~\cite{ganin2015unsupervised}, and USPS~\cite{denker1988neural}, with variance on foreground shapes and background patterns. 
MNIST is used as the source domain containing 60,000 training images. 
We convert all the images to $32\times32$ resolution with RGB format in the experiment. 

\noindent\textbf{CIFAR-10-C.} 
\textit{CIFAR-10-C} dataset~\cite{hendrycks2019benchmarking} is the corrupted version from the original \textit{CIFAR-10}~\cite{krizhevsky2009learning} dataset, including 10 classes and totally 50,000 training images with $32\times32$ resolution. 
There are 4 categories of corruption including weather, blur, noise, and digital. 
For each category, the corruption level is marked from 1 (mildest) to 5 (severest). 

\begin{table}[!t]
\scriptsize
\centering
    \begin{tabular}{cccccc}
        \toprule
        Method & SVHN & M-MNIST & SYN & USPS & Avg. \\
        \midrule
        Source Only & 27.83 & 52.72 & 39.65 & 76.94 & 49.29 \\
        \midrule
        JiGen & 33.80 & 57.80 & 43.79 & 77.15 & 53.14 \\
        RSC & 31.04 & 46.62 & 34.81 & 64.42 & 44.22 \\
        MMLD & 26.41 & 51.51 & 38.33 & 75.04 & 47.82 \\
        ADA & 35.51 & 60.41 & 45.32 & 77.26 & 54.62 \\
        M-ADA & 42.55 & 67.94 & 48.95 & 78.53 & 59.49 \\
        ME-ADA & 42.56 & 63.27 & 50.39 & 81.04 & 59.32 \\
        MixStyle & 32.29 & 53.48 & 42.35 & 81.17 & 52.32 \\
        L2D & \underline{62.86} & \underline{87.30} & \underline{63.72} & \underline{83.97} & \underline{74.46} \\
        \midrule
        Ours & \textbf{67.48} & \textbf{90.75} & \textbf{69.40} & \textbf{87.64} & \textbf{78.82} \\
        \bottomrule
    \end{tabular}
    \vspace{-0.3cm}
    \caption{Comparisons of single DG accuracy (\%) on Digits. MNIST is used for training while the others are for evaluation.}
    \vspace{-0.6cm}
    \label{tab:digits}
\end{table}

\noindent\textbf{PACS.} 
\textit{PACS} dataset~\cite{li2017deeper} contains 9,991 images of 4 domains: photo, art painting, cartoon, and sketch with 7 classes. 
The cross-domain variance in style and deformation is considerable and the adopted resolution is $224\times224$, which makes it a more challenging benchmark. 

\noindent\textbf{KITTI and vKITTI:}
\textit{KITTI}~\cite{geiger2013vision} is an outdoor dataset with 42,382 images for automatic driving. 
In this paper, we use the test dataset for evaluation. 
The training domain is \textit{vKITTI} dataset~\cite{gaidon2016virtual} containing 21,260 frames with depth labels from the Unity game engine. 
All the images are resized to $640\times192$ resolution for training and evaluation.

\begin{table*}[!t]
\scriptsize
\centering
    \begin{tabular}{cc|c|ccccccc|c}
        \toprule
        \multicolumn{2}{c}{Settings} & Source Only & JiGen & RSC & MMLD & ADA & ME-ADA & MixStyle & L2D & Ours \\
        \midrule
        \multirow{4}{*}{Photo} & Art & 62.26 & 60.74 & 67.72 & 64.59 & 64.31 & 65.62 & 67.42 & \underline{68.07} & \textbf{72.12} \\
                               & Cartoon & 27.60 & 33.40 & 33.70 & 30.25 & 34.94 & \underline{36.95} & 36.34 & 34.43 & \textbf{55.03} \\
                               & Sketch & 29.35 & 43.96 & \underline{48.00} & 28.61 & 36.12 & 35.10 & 38.28 & 44.69 & \textbf{62.61} \\
                               \cmidrule{2-11}
                               & Avg. & 39.73 & 46.03 & \underline{49.81} & 41.15 & 45.12 & 45.89 & 47.35 & 49.06 & \textbf{63.25} \\
        \midrule
        \multirow{4}{*}{Art} & Photo & 96.29 & \underline{96.71} & 92.75 & 96.47 & 95.81 & 95.69 & \textbf{97.23} & 96.11 & 94.13 \\
                               & Cartoon & 61.01 & 58.40 & \underline{71.89} & 55.97 & 67.96 & 67.28 & 64.66 & 70.61 & \textbf{71.97} \\
                               & Sketch & 49.25 & 51.23 & \underline{69.43} & 41.46 & 68.26 & 65.31 & 54.32 & 65.08 & \textbf{74.09} \\
                               \cmidrule{2-11}
                               & Avg. & 68.85 & 68.78 & \underline{78.02} & 64.63 & 77.34 & 76.09 & 72.07 & 77.26 & \textbf{80.06} \\
        \midrule
        \multirow{4}{*}{Cartoon} & Photo & 85.27 & 85.57 & 85.33 & 85.33 & 85.99 & 84.49 & \underline{87.72} & 86.17 & \textbf{87.78} \\
                               & Art & 63.38 & 68.90 & 71.00 & 62.11 & 68.55 & 57.82 & 71.59 & \textbf{75.24} & \textbf{75.93} \\
                               & Sketch & 67.73 & 63.35 & 73.30 & 66.07 & 72.28 & 71.82 & 63.78 & \underline{73.40} & \textbf{75.87} \\
                               \cmidrule{2-11}
                               & Avg. & 72.13 & 72.60 & 76.54 & 71.17 & 75.61 & 74.71 & 74.36 & \underline{78.27} & \textbf{79.86} \\
        \midrule
        \multirow{4}{*}{Sketch} & Photo & 24.73 & 36.65 & 44.25 & 21.13 & 25.33 & 26.53 & 27.10 & \underline{48.63} & \textbf{58.80} \\
                               & Art & 22.61 & 28.61 & \underline{52.00} & 18.36 & 27.88 & 28.61 & 26.20 & 48.38 & \textbf{60.11} \\
                               & Cartoon & 41.13 & 41.30 & 61.86 & 34.04 & 58.70 & 52.89 & 52.07 & \underline{62.88} & \textbf{67.75} \\
                               \cmidrule{2-11}
                               & Avg. & 29.49 & 35.51 & 52.70 & 24.51 & 37.30 & 36.01 & 35.12 & \underline{53.40} & \textbf{62.22} \\
        \midrule
        \multicolumn{2}{c}{Avg.} & 52.55 & 55.73 & 64.27 & 50.37 & 58.84 & 58.18 & 57.23 & \underline{64.50} & \textbf{71.35} \\
        \bottomrule
    \end{tabular}
    \vspace{-0.3cm}
    \caption{Comparisons of single DG accuracy (\%) on the PACS dataset. The first column indicates the training domain while the second column indicates the unseen test domain. Results under the ResNet-18 backbone are reported. Best performances in comparisons are highlighted in \textbf{bold} and the second best ones are marked with \underline{underlines}.}
    \vspace{-0.4cm}
    \label{tab:pacs_resnet}
\end{table*}

\subsection{Comparison with Sate-of-the-arts}
We mainly compare StyDeSty with 8 state-of-the-art single DG methods, including Jiasaw-puzzle based JiGen~\cite{carlucci2019domain}, self-challenging based RSC~\cite{huang2020self}, clustering based MMLD~\cite{matsuura2020domain}, adversarial augmentation based ADA~\cite{volpi2018generalizing}, M-ADA~\cite{qiao2020learning}, ME-ADA~\cite{zhao2020maximum}, and style enhancement based MixStyle~\cite{zhou2021domain} and L2D~\cite{wang2021learning}, as well as the Source Only baseline. 
All the comparisons are conducted using the same datasets and backbone networks. 
For our method, by default, the batch size is set as $64$ and the optimizer is SGD. 
The optimizer for $F$ and $H$ uses a $0.0005$ weight decay and $0.9$ momentum with the Nesterov mode~\cite{nesterov1983method}. 
Learning rates for $F$, $H$, and $G$ are $0.001$, $0.001$, and $0.005$. 
The times of inner iteration in \cref{alg:train} are all $1$ except that $T_H$ is $10$. 
The hyper parameters $\alpha$, $\lambda$, and $\beta$ are $0.1$, $1$, and $1$ respectively. 
As for the semantic perceptual network $M$ in \cref{eq:loss_G} and \cref{eq:loss_nas_G}, we directly use the task head $H$ itself as $M$ for all the classification tasks and features of the last hidden layer are adopted for $\mathcal{L}_{sem.}$. 
For the depth estimation problem, we load a fixed VGG19 model~\cite{simonyan2014very} pretrained on ImageNet~\cite{russakovsky2015imagenet} as $M$ and $\mathcal{L}_{sem.}$ would use features of the \texttt{ReLU-4\_1} layer.

\begin{table}[!t]
\scriptsize
\centering
    \begin{tabular}{ccccc}
        \toprule
        Method & Art & Cartoon & Sketch & Avg. \\
        \midrule
        w/o Style & 66.89	& 41.64 &	37.06 &	48.53 \\
        AutoAug & 70.80 &	44.50 &	50.09	& 55.13 \\
        DCGAN & \textbf{73.54}	& 47.01	& 49.50	& 56.68 \\
        w/o Destyle & 69.68 & 41.42 & 41.33 & 50.81 \\
        Separate Style Transfer & 70.51 &	53.33	& 58.39 &	60.74 \\
        w/o $\mathcal{L}_{align.}$ & 72.51 & 49.70 & 56.20 & 59.47 \\
        w/o $Percpt.$ & 69.53 & 47.35 & 52.25 & 56.38 \\
        end-to-end & 68.85 & 45.65 & 55.41 & 56.63 \\
        w/o Adv. & 71.58 &	53.63 &	59.63 &	61.61 \\
        \midrule
        Ours & 72.12 & \textbf{55.03} & \textbf{62.61} & \textbf{63.25} \\
        \bottomrule
    \end{tabular}
    \vspace{-0.3cm}
    \caption{Ablation studies of single DG accuracy (\%) on PACS dataset. The photo domain is used for the training domain while the other three are for evaluation.}
    \vspace{-0.5cm}
    \label{tab:pacs}
\end{table}

\noindent\textbf{Comparisons on Digits.} 
On the Digits dataset, we follow previous works and adopt the 5-layer LeNet~\cite{lecun1998gradient} as the backbone. 
There are 6 candidate positions to insert the AdaIN layer which are positions after the first and second convolution layer, ReLU layer, and pooling layer. 
NAS selects the position after the first pooling layer. 

The model is trained on the MNIST and evaluated on the other four. 
Comparisons of results by different methods are shown in \cref{tab:digits}. 
The improvement over the state-of-the-art methods is consistent: $4.62\%$, $3.45\%$, $5.68\%$, and $3.67\%$ on the 4 test datasets, which outperforms the state-of-the-art method \tcr{L2D~\cite{wang2021learning}} by $4.36\%$ in average. 

\begin{figure*}[!t]
    \centering
    \includegraphics[width=\linewidth]{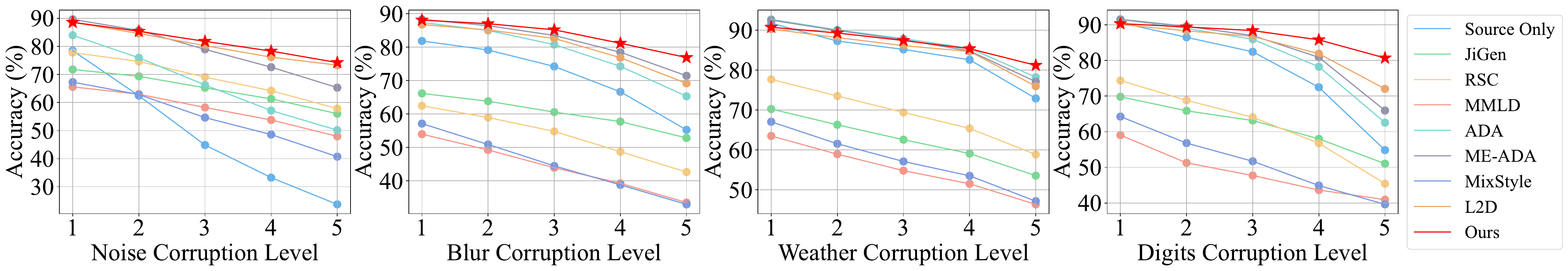}
    \vspace{-0.9cm}
    \caption{Relationships between accuracy results and corruption levels of four categories on CIFAR-10-C dataset. Our method demonstrates more robustness compared with other methods as the corruption increases.}
    \vspace{-0.5cm}
    \label{fig:corruption}
\end{figure*}

\begin{table}[!t]
\scriptsize
\centering
    \begin{tabular}{m{1.8cm}<{\centering}m{0.8cm}<{\centering}m{0.8cm}<{\centering}m{0.8cm}<{\centering}m{0.8cm}<{\centering}m{0.6cm}<{\centering}}
        \toprule
        Method & Photo & Art & Cartoon & Sketch & Avg. \\
        \midrule
        Source Only & 96.05 & 75.68 & 74.02 & 69.87 & 78.91 \\
        \midrule
        JiGen & \textbf{96.47} & 80.62 & 74.71 & 72.43 & 81.06 \\
        RSC & 93.95 & 82.81 & 79.74 & \textbf{83.51} & 85.00  \\
        MMLD & \underline{96.33} & 82.81 & 78.33 & 75.29 & 83.19 \\
        ADA & 95.63 & 82.81 & 78.33 & 75.29 & 83.02 \\
        ME-ADA & 95.33 & 77.88 & 78.58 & 78.07 & 82.47 \\
        MixStyle & 96.31 & 83.11 & 79.43 & 72.95 & 82.95 \\
        L2D & 95.15 & \underline{83.69} & \textbf{80.16} & 82.01 & \underline{85.25} \\
        \midrule
        Ours & 95.27 & \textbf{84.03} & \underline{79.86} & \underline{82.23} & \textbf{85.35} \\
        \bottomrule
    \end{tabular}
    \vspace{-0.3cm}
    \caption{Comparisons of leave-one-domain-out classification accuracy (\%) on PACS using ResNet-18 backbone. The column name indicates the test domain and the other three are used for training.}
    \vspace{-0.7cm}
    \label{tab:leave_one_out}
\end{table}

\noindent\textbf{Comparisons on CIFAR-10-C.} 
Following the common setting, we adopt WideResNet (16-4)~\cite{zagoruyko2016wide} as the backbone for the CIFAR-10-C dataset. 
There is one convolution layer followed by three residual blocks for this backbone, which means 
that the number of candidate positions for the AdaIN layer is 4. 
The optimal position indicated by the NAS algorithm is after the first residual block. 
The batch size used is $128$ and learning rates for $F$, $H$, and $G$ are $0.1$, $0.1$, and $0.001$ respectively. 

In this experiment, the original CIFAR-10 dataset is used as the training domain and the corrupted images are used for evaluation. 
The accuracy results w.r.t the corruption level for each method \tcr{are plotted as \cref{fig:corruption}}, which demonstrates that the model by our method can resist image corruption most robustly. 
On the severest corruption level 5, our method outperforms others by $0.94\%$, $5.46\%$, $2.96\%$, and $8.67\%$ for the four corruption categories respectively, and makes an average improvement of $5.65\%$.

\noindent\textbf{Comparisons on PACS.} 
We adopt ResNet-18~\cite{he2016deep} backbone on the PACS dataset. 
As a convention, a pre-trained checkpoint on ImageNet dataset~\cite{russakovsky2015imagenet} is loaded for initialization. 
We consider positions after the 4 main blocks as candidate AdaIN positions and the solution by the NAS algorithm is after the 2nd block. 

Following previous arts, the results of using each of the four domains for training respectively, and the other three for evaluation are reported in \cref{tab:pacs_resnet}. 
In almost all cases, our method performs significantly better than previous state-of-the-art ones, especially for scenarios where the domain shift is dramatic, like $14.61\%$ improvement when generalizing from the photo to the sketch domain. 
On average, our method outperforms others by $13.44\%$ when using the photo domain for training and $6.85\%$ overall by averaging the four training domains. 
Readers can refer to the supplementary material for results of more different architectures of backbone networks.

We also conduct experiments under the leave-one-out setting of general DG, to use three domains for training and the remaining one for evaluation, by mixing the data of three domains as one domain. 
ResNet-18 is used as the backbone and the selected AdaIN positions are all between the 1st and 2nd residual blocks in the 4 cases. 
Results in \cref{tab:leave_one_out} prove that our method can produce at least comparable performance without any constraint on label space, which indicates the versatility of the proposed method.

\begin{table}[!t]
\scriptsize
\centering
    \begin{tabular}{cccccc}
        \toprule
        Method & Photo & Art & Cartoon & Sketch & Avg. \\
        \midrule
        Ours & 63.25	&  80.06   &  79.86    &	62.22 &	71.35 \\
        \midrule
        \cite{lv2022causality} & 46.21 &	75.64 &	78.29 &	58.44 &	64.65 \\
        \cite{lv2022causality}+Ours & 54.36 &	79.33 &	78.99 &	60.77 &	68.36 \\
        \midrule
        \cite{chen2023meta} & 57.99 &	76.18 &	77.91 &	58.11 &	67.55 \\
        \cite{chen2023meta}+Ours & 64.36 &	78.46 &	79.62	& 59.63	& 70.52 \\
        \midrule
        \cite{choi2023progressive} & 62.89  &	76.98 &	78.54 &	57.11 &	68.88 \\
        \cite{choi2023progressive}+Ours & \textbf{67.98}    &	\textbf{81.82}	&  \textbf{80.80}  &	\textbf{63.15}    &	\textbf{73.44} \\
        \bottomrule
    \end{tabular}
    \vspace{-0.3cm}
    \caption{Single DG accuracy (\%) on PACS dataset when our method is built on state-of-the-art ones as a plug-and-play component. The column name indicates the training domain, and the other three are used for training. The average performance over the three test domains is reported.}
    \vspace{-0.5cm}
    \label{tab:add}
\end{table}

\textbf{Comparisons on KITTI:} 
In addition to the above classification tasks, we conduct experiments on a regression problem: monocular depth estimation on the KITTI dataset. 
The backbone network is a 4-level UNet-like~\cite{ronneberger2015u} architecture following ~\cite{zhao2019geometry}. 
For our method, we insert AdaIN layers to the upper three skip connection structures as well as the bottom level. 
Adam~\cite{kingma2014adam} is used as the optimizer with a learning rate of $0.0001$. 

In this part, we mainly compare StyDeSty in this paper with those methods without constrain on the label space, including MixStyle~\cite{zhou2021domain} and a variant of L2D~\cite{wang2021learning} by removing the class-conditional terms that are incompatible with this task, denoted as v-L2D. 
Using vKITTI as the source domain, evaluation results on the test KITTI dataset are shown in \cref{tab:depth}. 
Through all the metrics, we can observe that our method generalizes to the unseen target domain best by learning style-invariant feature representations, which demonstrates the versatility and superiority of StyDeSty. 

\subsection{Empirical Analysis}

\begin{table}[!t]
\scriptsize
\setlength{\tabcolsep}{0.8pt}
\centering
    \begin{tabular}{cccccccc}
        \toprule
        \multicolumn{1}{c}{\multirow{2}{*}{Method}} & \multicolumn{3}{c}{Higher is better} & \multicolumn{4}{c}{Lower is better} \\
        \cmidrule(r){2-4}\cmidrule(r){5-8}
        & $\delta<1.25$ & $\delta<1.25^2$ & $\delta<1.25^3$ & Abs Rel & Squa Rel & RMSE & $\rm{RMSE}_{log}$ \\
        \midrule
        Source Only & 0.642 & 0.861 & 0.944 & 0.236 & 2.171 & 7.063 & 0.315 \\
        \midrule
        MixStyle & 0.701 & 0.887 & 0.952 & 0.216 & 2.155 & 6.895 & \underline{0.291} \\
        v-L2D  & \underline{0.708} & \underline{0.892} & \underline{0.954} & \underline{0.211} & \underline{2.103} & \underline{6.794} & \underline{0.291} \\
        \midrule
        Ours & \textbf{0.739} & \textbf{0.905} & \textbf{0.958} & \textbf{0.197} & \textbf{2.054} & \textbf{6.684} & \textbf{0.276} \\
        \bottomrule
    \end{tabular}
    \vspace{-0.45cm}
    \caption{Comparisons of monocular depth estimation on KITTI dataset. The vKITTI dataset extracted from a game engine is used for the training domain and the KITTI test dataset is for evaluation. Best performances in comparisons are highlighted in \textbf{bold} and the second best ones are marked with \underline{underlines}.}
    \vspace{-0.7cm}
    \label{tab:depth}
\end{table}

\noindent\textbf{Ablation Study.} 
To validate the effectiveness of some key designs in our StyDeSty framework, we conduct ablations on the PACS dataset as shown in \cref{tab:pacs}. 
We first study the effect of the stylization module, which intuitively diversifies the source domain and tells the model what information can be potentially variable in the inference time. 
Without this module, the model is unaware of domain-invariant features, leading to the misalignment of features that should remain distinct. 
As shown in the first three settings of \cref{tab:pacs}, we try both removing the data augmentation module and replacing it with augmentations other than the stylization in this paper, like the widely used AutoAug~\cite{DBLP:journals/corr/abs-1805-09501} a vanilla DCGAN~\cite{DBLP:journals/corr/RadfordMC15} generator. 
The results are measured on PACS taking Photo as the training domain and the remaining ones as test domains. 
Their inferior performance compared with the default setting of StyDeSty verifies the effectiveness of the stylization module. 

Then, we delete the AdaIN layer for destylization (w/o Destyle) and find that the performance would drop significantly, which demonstrates that explicit feature alignment contributes to the generalization ability a lot. 
We also try replacing the destylization module with a pre-trained style transfer module~\cite{huang2017arbitrary}, which is unaware of the downstream task. 
The performance gap indicates the effectiveness of the task-aware destylization in our method. 
Plus, only aligning the second-order statistics does not make the model aware of invariant features without the metric $\mathcal{L}_{align.}$ in \cref{eq:loss_F}(w/o $\mathcal{L}_{align.}$) and the model would produce inferior performance compared with that of the full model. 
Moreover, if only the $L2$ loss between normalized features is considered in $\mathcal{L}_{align.}$ without the perceptual term (w/o $Percpt.$), the performance can become even worse. 
However, it is non-trivial to add the perceptual term in the task-specific feature space of $H_{-1}$. 
If the network is trained in an end-to-end manner with $F$ and $H$ being updated at the same time (end-to-end), which means that the task head would also be affected by the perceptual alignment loss, the ability of the destylization module to learn a unified distribution would be weakened and the performance is also unsatisfactory. 
That is why StyDeSty uses a multi-stage training strategy and achieves the best performance, \tcr{which drives} both explicit distribution alignment and an appropriate constraint on aligned features. 

We finally make the stylization module a random style augmenter in each iteration instead of playing against the destylization module adversarially. 
The performance drop demonstrates the effectiveness of their interplay. 

\begin{figure}[t]
    \centering
    \includegraphics[width=\linewidth]{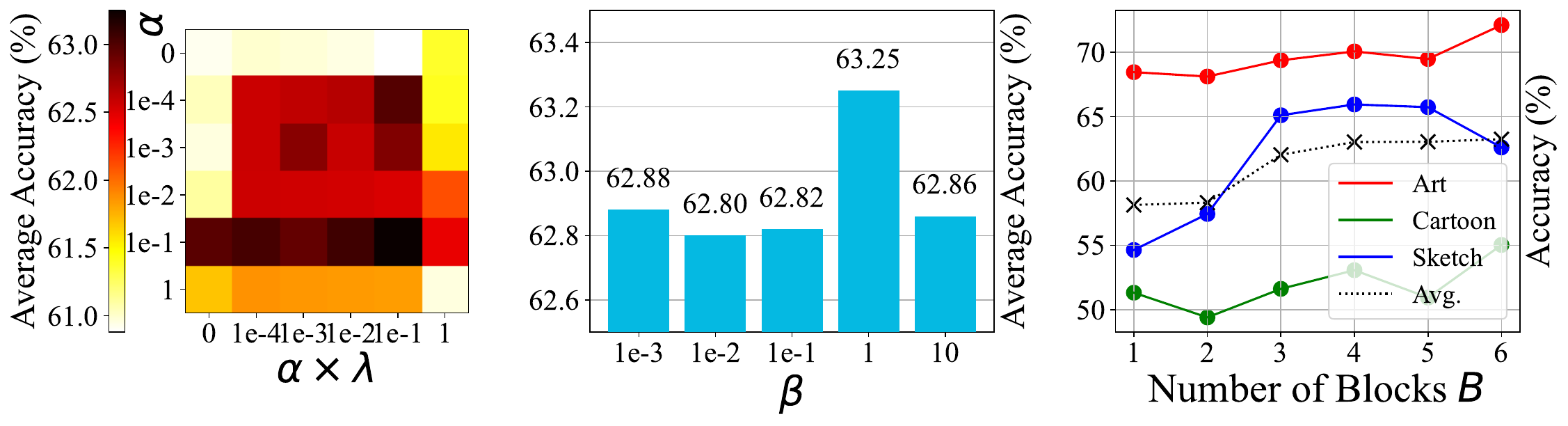}
    \vspace{-0.9cm}
    \caption{Sensitivity analysis for loss weights: $\alpha$, $\beta$, $\lambda$, and $B$.}
    \vspace{-0.7cm}
    \label{fig:hyper}
\end{figure}

\noindent\textbf{Complementarity with Other Methods.} 
The destylization mechanism can also serve as a plug-and-play component to improve the performance of other state-of-the-art methods. 
We apply the destylization layer and the corresponding position found by the NAS algorithm to the backbones of \cite{lv2022causality}, \cite{chen2023meta}, and \cite{choi2023progressive}, respectively. 
The results on PACS are shown in \cref{tab:add}, where the single DG results are shown. 
The column name indicates the training domain, and the other three are used for training. 
We report the average performance over the three test domains. 
The results indicate that our method can improve other single DG methods as a general enhancer. 
When applied on \cite{choi2023progressive}, it achieves even better performance than the original StyDeSty framework in the default setting of this paper. 

\noindent\textbf{Sensitivity \jx{Study} for Hyper-parameters.} 
We conduct analysis for  loss weights: $\alpha$, $\beta$, and $\lambda$ on the PACS dataset. 
As shown in \cref{eq:loss_F}, the overall weights of the $L2$ distance and the perceptual term are $\alpha$ and $\alpha\times\lambda$ respectively, whose sensitivity is analyzed in \cref{fig:hyper}(left). 
The sensitivity of $\beta$, the weight of the semantic consistency in \cref{eq:loss_G}, is analyzed in \cref{fig:hyper}(middle). 
We observe performance variation up to $2\%$ across different values, which reveals the robustness of our method to the various hyper-parameters. 
We also study the parameter $B$ of \cref{eq:2} in \cref{fig:hyper}(right) and find that the performance is insensitive to $B$ if there are sufficient augmentation modules.  

\begin{figure}[t]
    \centering
    \includegraphics[width=\linewidth]{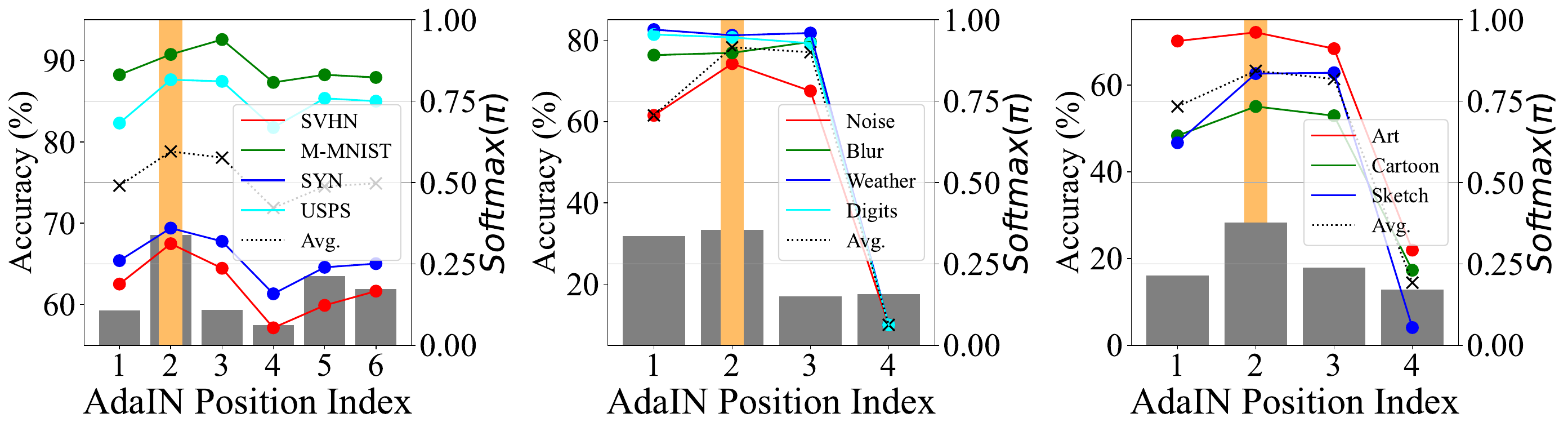}
    \vspace{-0.8cm}
    \caption{Accuracy by conducting destylization at different positions in a backbone network for 3 different settings: LeNet-5 on Digits, WideResNet (16-4) on CIFAR-10-C, and ResNet-18 on PACS. Positions selected by the NAS algorithm are highlighted in orange. We also visualize the corresponding probabilistic scores ${\rm Softmax}(\pi)$ of \cref{eq:nas} in grey.}
    \vspace{-0.7cm}
    \label{fig:nas}
\end{figure}

\noindent\textbf{NAS Algorithm.} 
In this part, we experiment with all the candidate positions for inserting the destylization AdaIN layer, to rationalize the final position selected by the NAS algorithm. 
The results of using LeNet-5 on the Digits dataset, WideResNet (16-4) on the CIFAR-10-C dataset, and ResNet-18 on the PACS dataset are shown as the three plots in \cref{fig:nas} respectively. 
The selected position is highlighted in orange. 
\lsh{The selected positions are stable when the NAS algorithm is \jx{executed} multiple times.}
In all settings, our NAS algorithm can find the optimal position in a backbone network to conduct destylization, in the sense of average accuracy over test domains, which proves its general effectiveness for different datasets and backbone networks. 
More analysis of the NAS algorithm can be found in the supplementary material. 

\section{Conclusions}
In this paper, we propose a simple yet effective 
approach for single DG, termed StyDeSty, 
by introducing the stylization and destylization mechanism. 
The stylization module aims to generate diversely stylized samples,
while the destylization module learns to unify and align the feature distributions. 
These two designs are co-optimized by a min-max game 
with a NAS-based method seeking for an optimal position for destylization. 
StyDeSty is a versatile framework that not only works for classification tasks but is also readily applicable to regression problems.
Extensive experiments on multiple benchmarks demonstrate that
StyDeSty significantly outperforms the state-of-the-art methods by up to $13.44\%$ in terms of classification accuracy. 


\section*{Impact Statement}

Single domain generalization is a critical area of research in machine learning that aims to enhance the robustness and adaptability of models trained on a single domain when applied to diverse and unseen domains. The impact of this research lies in its potential to significantly improve the deployment of AI systems in real-world scenarios where collecting comprehensive and diverse training data is impractical. By advancing techniques that enable models to generalize from a limited dataset, this work can lead to more reliable and versatile AI applications across various industries, from healthcare and autonomous driving to finance and beyond. Ultimately, single domain generalization fosters the development of more resilient AI systems, contributing to safer and more efficient technological solutions.

\section*{Acknowledgments}
This project is supported by the Singapore Ministry of Education Academic Research Fund Tier 1 (WBS: A-8001229-00-00), a project titled “Towards Robust Single Domain Generalization in Deep Learning”.

\bibliography{StyDeSty}
\bibliographystyle{icml2024}

\clearpage
\appendix

In this appendix, we provide more discussion with related works, more analysis, additional details, and more comparison results of the proposed StyDeSty framework for single domain generalization (single DG). 
First, we summarize related works in a table as a supplement to the related work section of the main paper. 
Then, we provide some qualitative examples to illustrate the motivation of the proposed method as a supplement to the main paper. 
We will also give more details on the implementation of the stylization module $G$ and some loss functions. 
Finally, we conduct more experiments to demonstrate and analyze the performance of our method, including results on more settings and benchmarks, and comparisons on the monocular depth estimation task.

\begin{table*}[!t]
    \centering
    \scriptsize
    \begin{tabular}{ccccc}
        \textbf{Method} & \textbf{Venue} & \textbf{Key Words} & \textbf{Alignment Loss} & \textbf{Explicit Alignment} \\
        \midrule
        JiGen~\cite{carlucci2019domain} & CVPR 2019 & Jigsaw Puzzles & No & No \\
        RSC~\cite{huang2020self} & ECCV 2020 & Self-Challenging & No & No \\
        MMLD~\cite{matsuura2020domain} & AAAI 2020 & Adversarial Augmentation & Yes & No \\
        ADA~\cite{volpi2018generalizing} & NeurIPS 2018 & Adversarial Augmentation & No & No \\
        M-ADA~\cite{qiao2020learning} & CVPR 2020 & Adversarial Augmentation & No & No \\
        ME-ADA~\cite{zhao2020domain} & NeurIPS 2020 & Adversarial Augmentation & Yes & No \\
        MixStyle~\cite{zhou2021domain} & ICLR 2021 & Style Mix-Up & No & No \\
        PDEN~\cite{li2021progressive} & CVPR 2021 & Stylization & Yes & No \\
        L2D~\cite{wang2021learning} & ICCV 2021 & Stylization & Yes & No \\
        MetaCNN~\cite{wan2022meta} & CVPR 2022 & Meta Feature Learning & No & No \\
        CIRL~\cite{lv2022causality} & CVPR 2022 & Causality & Yes & No \\
        ABA~\cite{cheng2023adversarial} & CVPR 2023 & Adversarial Augmentation & Yes & No \\
        MAD~\cite{qu2023modality} & CVPR 2023 & Debiasing & No & No \\
        Meta-Causal~\cite{chen2023meta} & CVPR 2023 & Causality & Yes & No \\
        ProRandConv~\cite{choi2023progressive} & CVPR 2023 & Augmentation & No & No \\
        \midrule
        Ours & ICML 2024 & Stylization and Destylization & Yes & Yes \\
        \hline
    \end{tabular}
    \caption{Summary of related works on single domain generalization methods.}
    \vspace{-0.5cm}
    \label{tab:sum}
\end{table*}

\section{Summary of Related Works}
We summarize the related works of single domain generalization methods in \cref{tab:sum}, focusing on method keywords, alignment loss, and explicit alignment. 
With the regularization of both \textbf{alignment loss} and \textbf{explicit alignment} in destylization, our method achieves superior single-domain generalization performance. 

\section{Motivation}
Here, as a supplement to \cref{fig:pre_exp} of the main paper, we provide further qualitative analysis to the three key questions: \emph{why, how, and where to destyle in single DG?}

\begin{figure}[t]
    \centering
    \includegraphics[width=\linewidth]{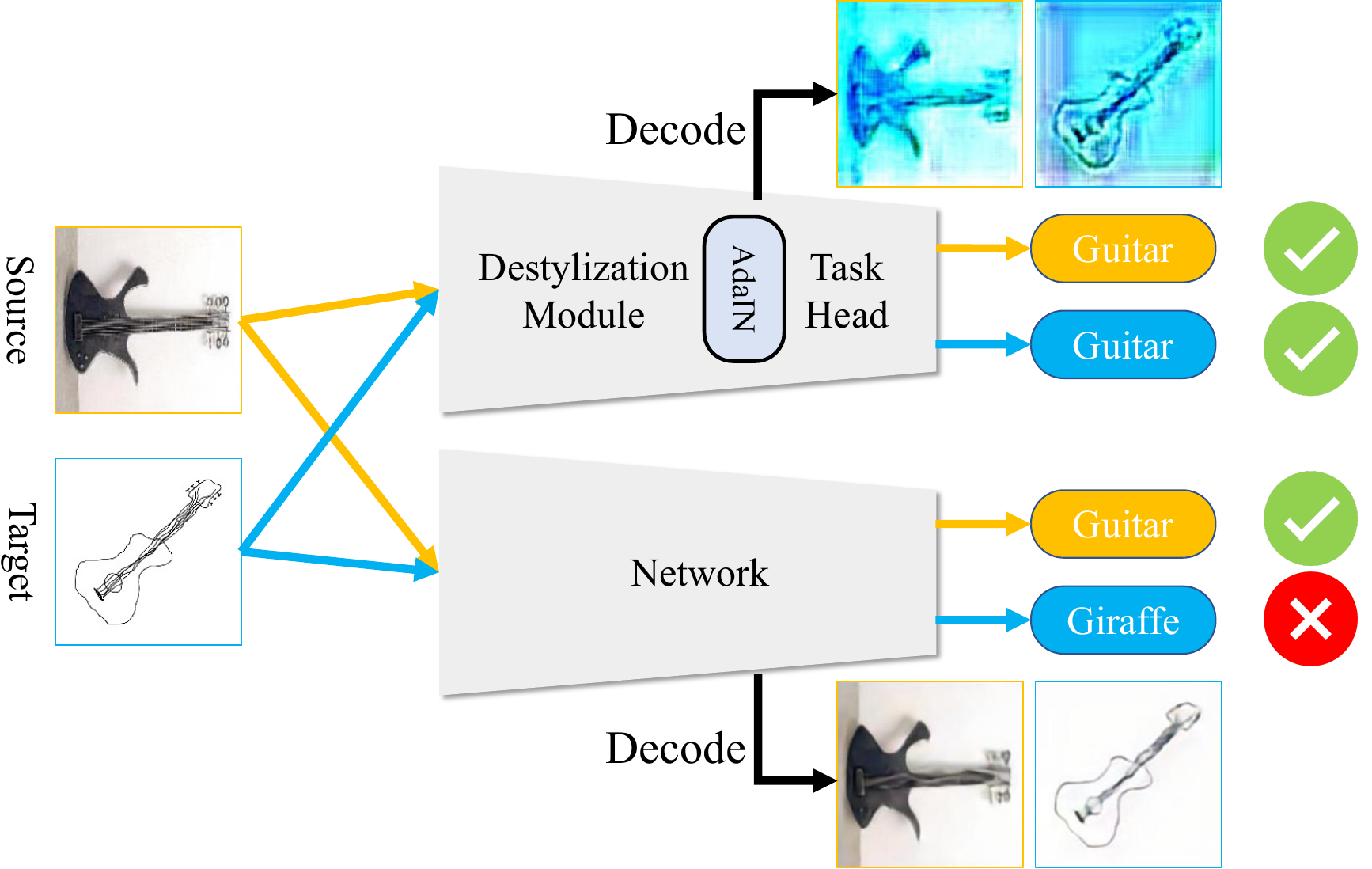}
    \vspace{-0.9cm}
    \caption{The explicit destylization mechanism unifies styles of the source domain and any unseen target domains, which benefits the performance of the downstream task head.}
    \label{fig:why}
    \vspace{-0.4cm}
\end{figure}

\begin{figure}[t]
    \centering
    \includegraphics[width=\linewidth]{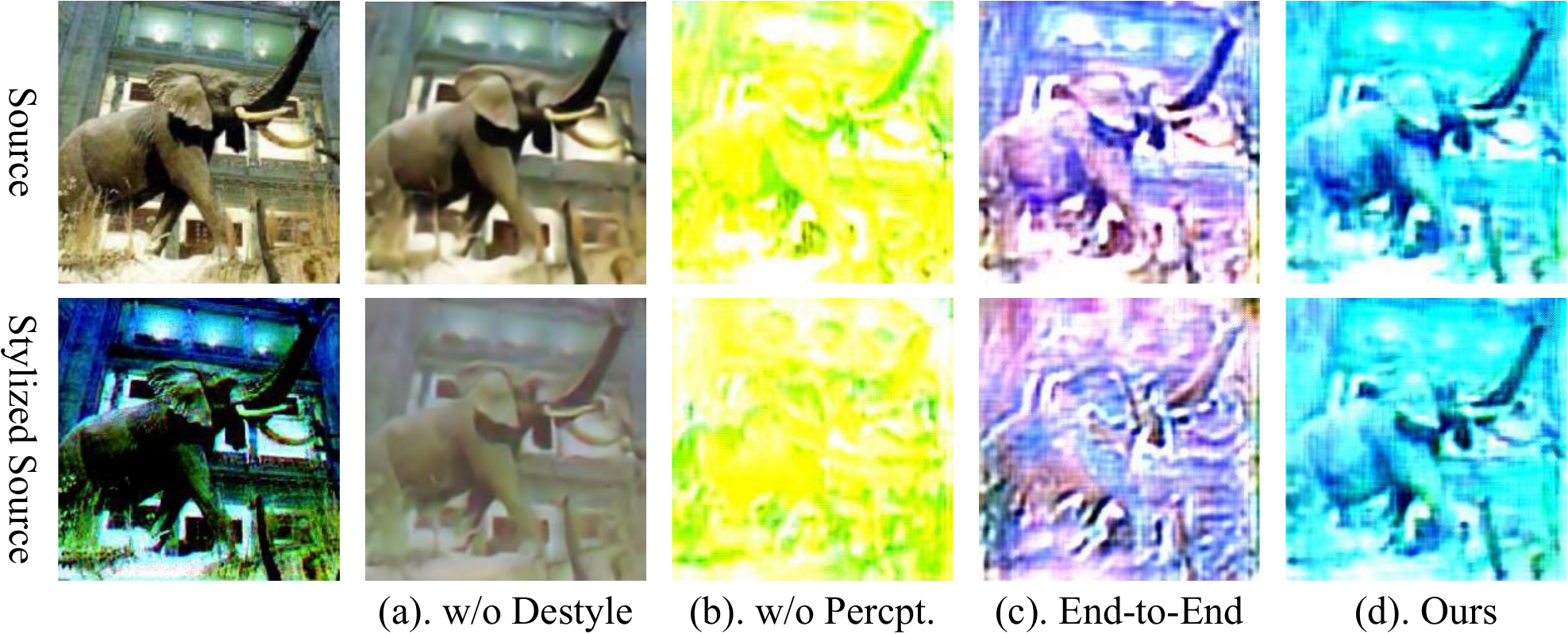}
    \vspace{-0.8cm}
    \caption{Qualitative results by different fashions of destylization.}
    \label{fig:how}
    \vspace{-0.4cm}
\end{figure}

\noindent\textbf{Why to Destyle?} 
Intuitively, the destylization module in this paper aims to transfer any unseen styles in the test time to the one most familiar to the task head. 
As shown in \cref{fig:why}(top), we decode features after the destylization AdaIN layer to the image space with a pre-trained decoder and find that styles of the source domain (photo) and the unseen target domain (sketch) are aligned, which benefits the following task head. 
By contrast, in \cref{fig:why}(bottom), without explicit destylization, the network is less robust to domain shift and results in inferior performance. 
There is another example in the plot (a) of \cref{fig:how} to illustrate this effect. 

\noindent\textbf{How to Destyle?} 
In this paper, there are two metrics to measure the effectiveness of destylization: the element-wise feature distance and the task-perceptual term measured in the space of the task head. 
On the one hand, if only the former one is adopted, the destylization module would not realize what properties are important for the following task. 
As shown in plot (b) of  \cref{fig:how}, although the overall styles are aligned, the major semantic structure is destroyed, which harms the downstream classification. 
On the other hand, if the task head for measuring the perceptual loss is trained jointly with the destylization module, the task head would also contribute to the destylization, which weakens the ability of alignment for the destylization module. 
As shown in the plot (c), the alignment is not satisfactory enough compared with that in the plot (d). 

\begin{figure}[t]
    \centering
    \includegraphics[width=\linewidth]{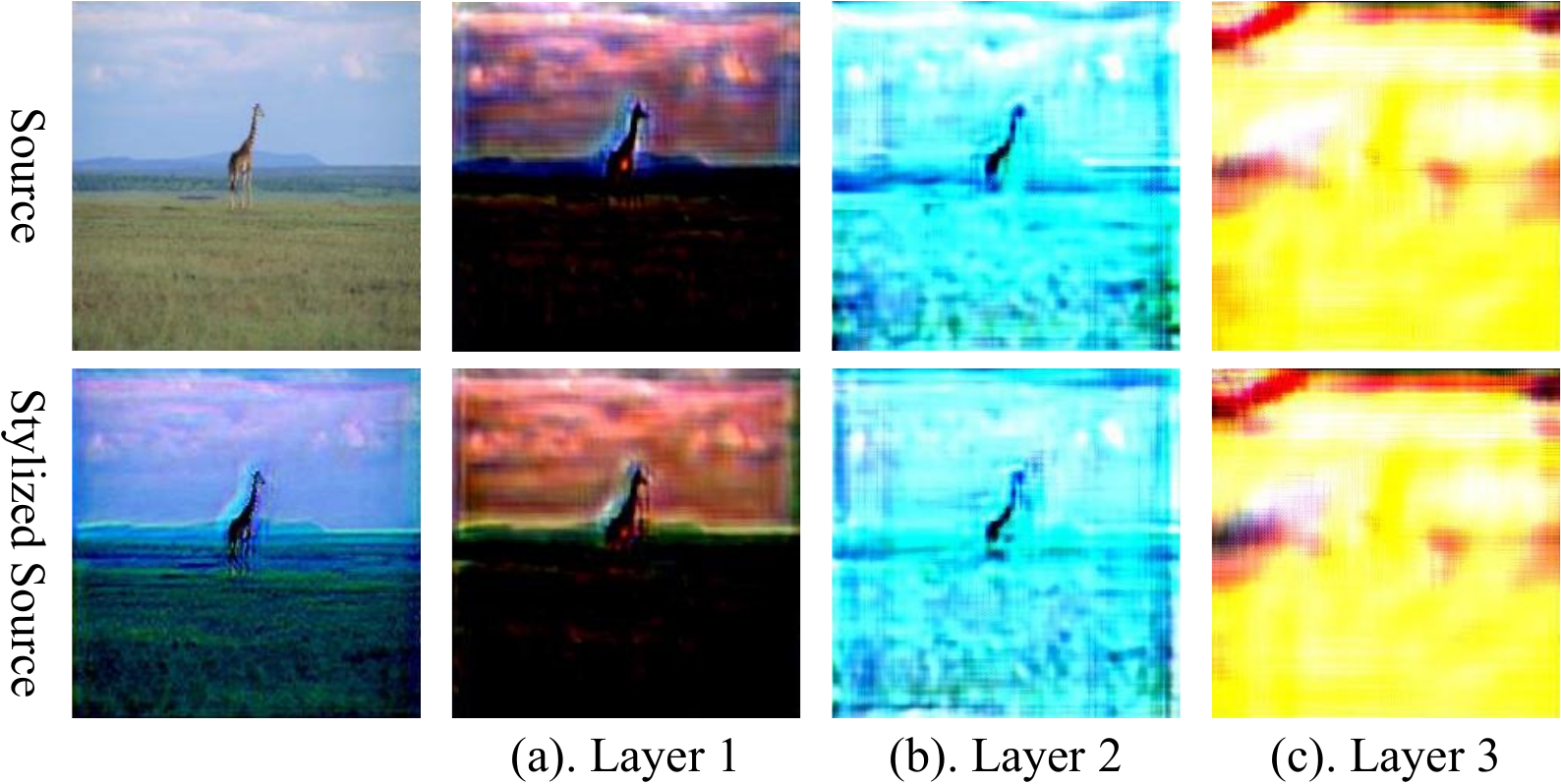}
    \vspace{-0.6cm}
    \caption{Qualitative results by destylization at different locations.}
    \label{fig:where}
    \vspace{-0.5cm}
\end{figure}

\noindent\textbf{Where to Destyle?} 
In the StyDeSty framework, the performance is sensitive to the location of the destylization layer AdaIN in a network. 
As shown in  \cref{fig:where}, as the location of destylization goes deeper, the alignment becomes more convenient but more discriminative information is lost. 
The trade-off between alignment and knowledge for the task motivates us to propose a NAS algorithm for better interaction among stylization, destylization, and the task module. 

\section{Model Details}

\noindent\textbf{Stylization Module.} 
As mentioned in the main paper, 
the stylization module $G$ consists of $B$ 
encoder-transformation-decoder blocks. 
All encoders take the form of  
a single convolution layer,
and all decoders take the form of a symmetric deconvolution layer. 
Encoders  project an image to
a $c$-dimension feature space $\mathbb{R}^{h\times w\times c}$,
and then the transformation layer learns to conduct affine transformation in this space. 
The affine parameters for some blocks have shape $h\times w\times c$ to account for  local distortions, and parameters for others have shape $1\times 1\times c$ to account for global distortions. 

In the experiments on small images with a $32\times32$ resolution, such as \textit{CIFAR-10-C}~\cite{hendrycks2019benchmarking,krizhevsky2009learning} and \textit{Digits}~\cite{lecun1998gradient,netzer2011reading,ganin2015unsupervised,denker1988neural} datasets, the stylization module uses $2$ blocks with local transformation for one and global transformation for another. 
The number of channels for both blocks is $3$ and the kernel size is $3$. 
For other classification tasks, the resolution of $224\times224$ is used,
and we use 4 blocks with local transformation and 2 with global transformation. 
The local transformation blocks have $3$ channels and kernel sizes are
$5$, $9$, $13$, and $17$. 
For the global transformation blocks, one has $3$ channels and a kernel size of $3$,
while the other has $64$ channels and a kernel size of $5$. 
The parameters of convolutional encoders and decoders are random for each iteration. 

\noindent\textbf{Metric of Perceptual Distance.} 
The feature alignment loss $\mathcal{L}_{align.}$ is a vital component of the objective for the destylization module $F$. 
It consists of two terms: $L2$ distance between two feature maps $F(x^\mathcal{S})$ and $F(x^\mathcal{T})$, and perceptual distance between $H_{-1}(F(x^\mathcal{S}))$ and $H_{-1}(F(x^\mathcal{T}))$, where $H_{-1}$ extracts features in the last hidden layer of the task head $H$, denoted as $h^\mathcal{S}$ and $h^\mathcal{T}$ for simplicity. 

To measure the distance between $h^\mathcal{S}$ and $h^\mathcal{T}$, 
we adopt the following negative log-likelihood~\cite{cheng2020club}:
$$
    -\frac{1}{n}\sum_{i=1}^n\log q_\theta(h_i^\mathcal{T}|h_i^\mathcal{S}),
$$
which employs a neural network parameterized by $\theta$ for variational inference. 
Please refer to \cite{cheng2020club} for the details. 

\noindent\textbf{Semantic Consistency Constraint.} 
To prevent the stylization module from destroying the semantic structures of an image and even generating meaningless results, a semantic consistency constraint that incorporates a semantic perceptron $M$ is included as one supervision signal for this module. 
Specifically, for the classification task, since the task itself is for semantic understanding, we directly adopt the task network, including the destylization module and the task head without the final linear layer, as the perceptron $M$. 
For tasks not directly related to semantic understanding like depth estimation, we introduce a pre-trained VGG19 encoder~\cite{simonyan2014very} as $M$, the semantic loss is measured on 5 layers: \texttt{ReLU-x\_1} for $1\leq x\leq5$.

\begin{figure*}[t]
    \centering
    \includegraphics[width=\linewidth]{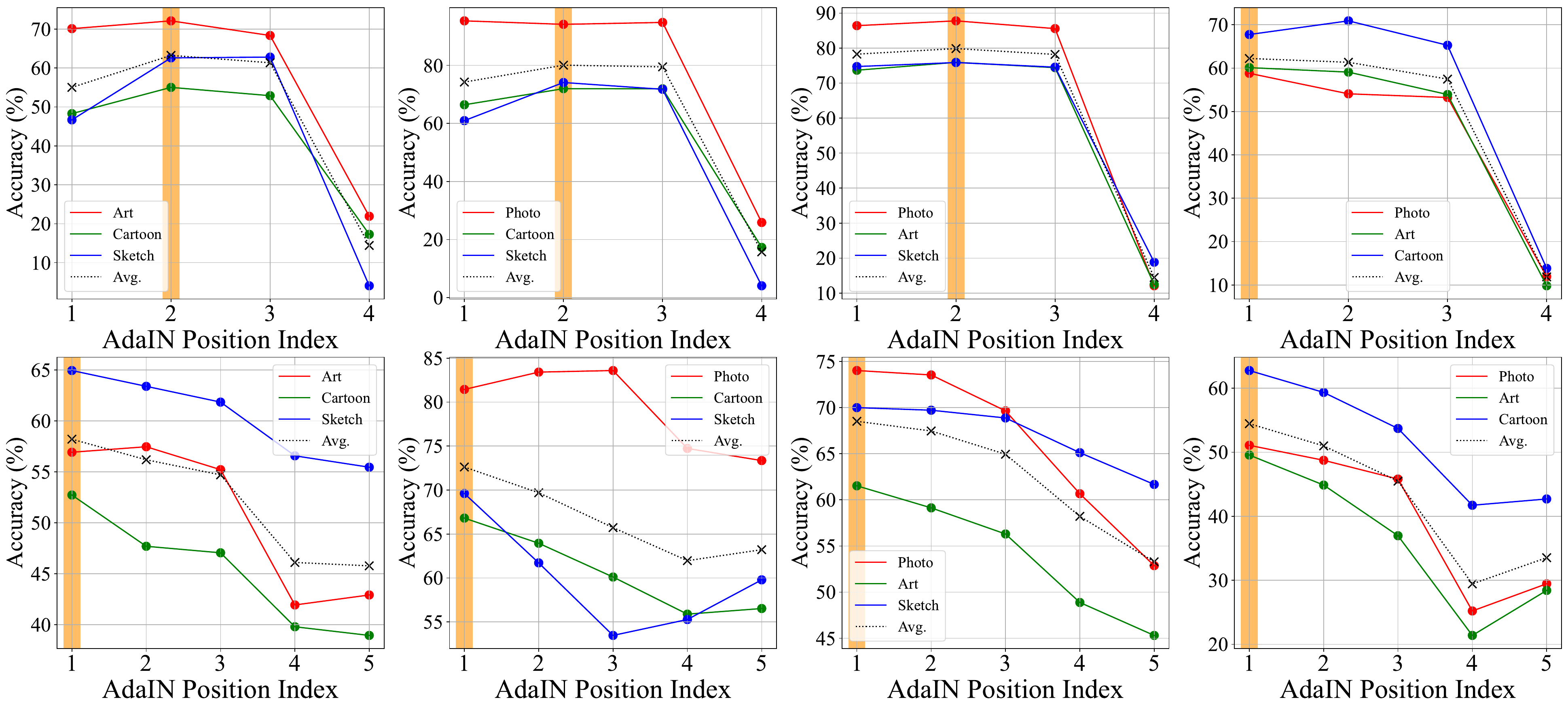}
    \vspace{-0.6cm}
    \caption{Accuracy results by conducting destylization at different positions in a backbone network for 4 training domains (photo, art painting, cartoon, and sketch for each column respectively) and 2 backbones (ResNet-18 and AlexNet for each row respectively) on PACS dataset. Positions selected by the NAS algorithm are highlighted in orange.}
    \label{fig:nas_supp}
    \vspace{-0.2cm}
\end{figure*}

\begin{table*}[!t]
\centering
\scriptsize
    \begin{tabular}{cc|c|ccccccc|c}
        \toprule
        \multicolumn{2}{c}{Settings} & Source Only & JiGen & RSC & MMLD & ADA & ME-ADA & MixStyle & L2D & Ours \\
        \midrule
        \multirow{4}{*}{Photo} & Art & 48.19 & 56.10 & 52.88 & \textbf{58.64} & 53.61 & 51.76 & 51.06 & \underline{58.45} & 56.93 \\
                               & Cartoon & 43.30 & 43.52 & 38.69 & \underline{49.87} & 45.44 & 44.67 & 41.42 & 49.74 & \textbf{52.73} \\
                               & Sketch & 51.31 & 52.46 & 48.69 & 50.24 & 49.02 & 50.62 & 48.06 & \underline{55.82} & \textbf{64.95} \\
                               \cmidrule{2-11}
                               & Avg. & 47.60 & 50.69 & 46.85 & 52.92 & 49.36 & 49.02 & 46.85 & \underline{54.67} & \textbf{58.21} \\
        \midrule
        \multirow{4}{*}{Art} & Photo & 80.24 & 84.07 & 85.21 & \textbf{90.36} & 81.44 & 81.38 & 81.42 & \underline{87.43} & 81.44 \\
                               & Cartoon & 61.05 & 65.23 & 60.54 & 60.49 & 61.48 & 59.68 & 58.81 & \textbf{69.03} & \underline{66.81} \\
                               & Sketch & 57.80 & 62.10 & 58.74 & 51.77 & 59.94 & 58.84 & 56.56 & \underline{66.38} & \textbf{69.61} \\
                               \cmidrule{2-11}
                               & Avg. & 66.36 & 70.47 & 68.16 & 67.54 & 67.62 & 66.64 & 65.60 & \textbf{74.28} & \underline{72.62} \\
        \midrule
        \multirow{4}{*}{Cartoon} & Photo & 64.91 & \underline{79.58} & 73.41 & \textbf{85.57} & 68.86 & 68.50 & 66.74 & 76.95 & 74.01 \\
                               & Art & 49.07 & 58.25 & 53.65 & 61.72 & 51.86 & 53.17 & 50.53 & \textbf{62.45} & \underline{61.52} \\
                               & Sketch & 58.74 & 64.49 & 63.76 & 61.82 & 58.56 & 56.53 & 57.44 & \underline{67.07} & \textbf{69.99} \\
                               \cmidrule{2-11}
                               & Avg. & 57.88 & 67.44 & 63.58 & \textbf{69.70} & 59.76 & 59.40 & 58.24 & \underline{68.82} & 68.51 \\
        \midrule
        \multirow{4}{*}{Sketch} & Photo & 39.88 & 48.74 & \textbf{53.71} & \underline{53.05} & 38.02 & 38.26 & 41.01 & 46.17 & 51.08 \\
                               & Art & 31.84 & 37.60 & 39.94 & \underline{43.51} & 32.08 & 32.37 & 34.53 & 35.50 & \textbf{49.56} \\
                               & Cartoon & 52.18 & 54.39 & 54.52 & \underline{61.43} & 54.95 & 55.38 & 55.21 & 57.98 & \textbf{62.76} \\
                               \cmidrule{2-11}
                               & Avg. & 41.30 & 46.91 & 49.39 & \underline{53.38} & 41.68 & 42.00 & 43.58 & 46.55 & \textbf{54.46} \\
        \midrule
        \multicolumn{2}{c}{Avg.} & 53.29 & 58.88 & 57.00 & 60.89 & 54.61 & 54.27 & 53.57 & \underline{61.08} & \textbf{63.45} \\
        \bottomrule
    \end{tabular}
    \caption{Comparisons of single DG accuracy (\%) on PACS dataset. The first column indicates the training domain while the second column indicates the unseen test domain. Results under the AlexNet backbone are reported. Best performances in comparisons are highlighted in \textbf{bold} and the second best ones are marked with \underline{underlines}.}
    \vspace{-0.5cm}
    \label{tab:pacs_alexnet}
\end{table*} 

\begin{table*}[!t]
\centering
\scriptsize
    \begin{tabular}{cc|c|ccccccc|c}
        \toprule
        \multicolumn{2}{c}{Settings} & Source Only & JiGen & RSC & MMLD & ADA & ME-ADA & MixStyle & L2D & Ours \\
        \midrule
        \multirow{4}{*}{Photo} & Art & 64.75 & 47.85 & 60.11 & 57.13 & 62.30 & 63.62 & 58.60 & \underline{64.84} & \textbf{67.53} \\
                               & Cartoon & 33.15 & 27.99 & 35.45 & 22.44 & 35.71 & 35.84 & 19.49 & \underline{46.12} & \textbf{50.00} \\
                               & Sketch & 29.12 & 26.80 & 40.06 & 17.43 & 31.53 & 30.24 & 19.30 & \underline{53.19} & \textbf{61.92} \\
                               \cmidrule{2-11}
                               & Avg. & 42.34 & 34.21 & 45.21 & 32.33 & 43.18 & 43.23 & 32.46 & \underline{54.72} & \textbf{59.82} \\
        \midrule
        \multirow{4}{*}{Art} & Photo & 83.72 & 85.99 & 87.78 & 90.18 & \underline{94.25} & 91.86 & \textbf{95.22} & 92.22 & 84.97 \\
                               & Cartoon & 57.68 & 47.53 & 64.46 & 54.82 & 61.52 & 68.94 & 52.34 & \textbf{71.54} & \underline{69.75} \\
                               & Sketch & 42.73 & 35.81 & 60.72 & 46.81 & 52.51 & 48.59 & 35.62 & \underline{64.21} & \textbf{69.23} \\
                               \cmidrule{2-11}
                               & Avg. & 61.38 & 56.44 & 70.99 & 63.94 & 69.43 & 69.80 & 61.06 & \textbf{75.99} & \underline{74.65} \\
        \midrule
        \multirow{4}{*}{Cartoon} & Photo & 75.25 & 77.60 & 72.69 & 80.60 & 84.37 & 82.40 & \underline{84.74} & \textbf{88.32} & 84.13 \\
                               & Art & 63.28 & 56.49 & 57.56 & 57.28 & 62.84 & 61.96 & 58.78 & \textbf{72.80} & \underline{72.51} \\
                               & Sketch & 54.03 & 45.41 & 67.40 & 50.47 & 59.97 & 57.62 & 48.64 & \underline{64.88} & \textbf{70.37} \\
                               \cmidrule{2-11}
                               & Avg. & 64.19 & 59.84 & 65.89 & 62.78 & 69.05 & 67.33 & 64.05 & \underline{75.33} & \textbf{75.67} \\
        \midrule
        \multirow{4}{*}{Sketch} & Photo & 35.30 & 40.42 & 45.09 & 37.01 & 40.18 & 39.94 & 49.97 & \textbf{55.51} & \underline{53.95} \\
                               & Art & 37.74 & 38.13 & 50.78 & 38.33 & 40.43 & 38.87 & 38.16 & \underline{41.26} & \textbf{52.34} \\
                               & Cartoon & 57.64 & 37.46 & \underline{64.46} & 44.03 & 62.29 & 59.17 & 55.52 & 58.53 & \textbf{64.59} \\
                               \cmidrule{2-11}
                               & Avg. & 43.56 & 38.67 & 53.44 & 39.79 & 47.63 & 45.99 & 47.88 & 51.77 & \textbf{55.96} \\
        \midrule
        \multicolumn{2}{c}{Avg.} & 52.87 & 47.27 & 58.88 & 49.71 & 57.32 & 56.59 & 51.36 & \underline{64.45} & \textbf{66.53} \\
        \bottomrule
    \end{tabular}
    \vspace{-0.2cm}
    \caption{Comparisons of single DG accuracy (\%) on PACS dataset. The first column indicates the training domain while the second column indicates the unseen test domain. Results under the VGG11 backbone are reported. Best performances in comparisons are highlighted in \textbf{bold} and the second best ones are marked with \underline{underlines}.}
    \label{tab:pacs_vgg}
    \vspace{-0.2cm}
\end{table*}

\section{More Results}
\noindent\textbf{Full results on PACS dataset.} 
In this part, we provide full results of single domain generalization using the 4 domains in the PACS dataset~\cite{li2017deeper} one by one as the training domain. 
Methods for comparison are the same as those in the main paper, including JiGen~\cite{carlucci2019domain}, RSC~\cite{huang2020self}, MMLD~\cite{matsuura2020domain}, ADA~\cite{volpi2018generalizing}, ME-ADA~\cite{zhao2020maximum}, MixStyle~\cite{zhou2021domain}, and L2D~\cite{wang2021learning}, as well as the baseline method Source Only. 
The results under ResNet-18~\cite{he2016deep} and AlexNet~\cite{krizhevsky2012imagenet} backbones are shown in  \cref{tab:pacs_resnet} and  \cref{tab:pacs_alexnet} respectively.

Through the results, we can observe that our StyDeSty outperforms previous state-of-the-art methods in most cases, which is consistent with the conclusion in the main paper. 
Notably, our method is more robust compared with others when the domain shift is dramatic. 
For example, when the training or testing domain is Sketch, our method can produce consistent improvement. 
On average, our method achieves $6.85\%$ and $2.37\%$ promotion over the previous state of the arts. 

For ResNet18, NAS chooses the position after the second residual block when the training domain is photo, art painting, and cartoon and the position after the first residual block when the training domain is sketch, to insert the destylization layer. 
For the AlexNet backbone, the selected position is after the first convolution stage for all four training domains. 
To demonstrate the effectiveness of the NAS algorithm, we experiment with all the candidate positions for inserting the destylization AdaIN layer. 
The results of using the two backbone networks on the four training domains are shown as the eight plots in  \cref{fig:nas_supp} respectively, as a supplement to \cref{fig:nas} in the main paper. 
It proves that the NAS algorithm is competent to find an optimal position in a backbone network to conduct the destylization.

To further explore the capacity of the NAS algorithm, we also experiment with the VGG11~\cite{simonyan2014very} backbone on the PACS dataset. 
The full results are shown in  \cref{tab:pacs_vgg}. 
In this experiment, we choose positions after the 8 \texttt{ReLU} layers as candidate positions for the AdaIN layer. 
The NAS algorithm chooses \texttt{ReLU-1\_1} layer for art painting and sketch domain and \texttt{ReLU-2\_1} layer for photo and cartoon. 
We empirically find that when the number of candidates is larger than $10$, the convergence of the NAS algorithm would become difficult, \textit{e.g.}, if we select position after all the $20$ functionality layers in the feature extractor of VGG11 as candidates, the solution would sway among several adjacent layers. 
More advanced NAS algorithms such as coarse-to-fine strategies are necessary to handle larger backbone networks. 
Involving multiple destylization layers in a backbone network is also 
a promising future research direction.

We also conduct experiments on state-of-the-art network backbones like ConvNeXt~\cite{liu2022convnet} and SWIN~\cite{liu2021Swin}. 
The results of our method and the L2D~\cite{wang2021learning} baseline are shown in \cref{tab:pacs_more}. 
The models are trained on the Photo domain and evaluated on the Art Painting, Cartoon, and Sketch domains. 
The AdaIN layer is inserted after the second stage for ConvNeXt and after the first stage for SWIN. 
The results indicate that our method outperforms the baseline without explicit destylization significantly.

\begin{table}[!t]
\centering
\scriptsize
    \begin{tabular}{cccccc}
        \toprule
        Backbone & Method & A & C & S & Avg. \\
        \midrule
        \multirow{2}{*}{ConvNeXt} & L2D & 60.32 & 53.49 & 67.76 & 60.52 \\
        & Ours & \textbf{66.20} & \textbf{54.98} & \textbf{74.09} & \textbf{65.09} \\
        \midrule
        \multirow{2}{*}{SWIN} & L2D & \textbf{74.80} & 49.29 & 52.77 & 57.95 \\
        & Ours & 72.92 & \textbf{52.93} & \textbf{61.26} & \textbf{62.37} \\ 
        \bottomrule
    \end{tabular}
    \caption{Comparisons of single DG accuracy (\%) on PACS dataset. The models are trained in the Photo domain and evaluated on the Art Painting (A), Cartoon (C), and Sketch (S) domains. Results under ConvNeXt-T and SWIN-T backbones are reported.}
    \label{tab:pacs_more}
\end{table}


\balance

\noindent\textbf{Results on DomainNet Dataset.} 
To demonstrate the scalability of the proposed StyDeSty framework for single DG, we conduct experiments on DomainNet dataset~\cite{peng2019moment}, which consists of images from six distinct domains, including photos (real), clipart, infograph, painting, quickdraw, and sketch. 
There are 48k to 172k images in each domain and 600k in total that are categorized into 345 classes. 
We train the ResNet-18 models on the photos and evaluate them on the other 5 domains. 
The results of our model and the L2D~\cite{wang2021learning} baseline are shown in \cref{tab:domainnet}, where our methods yield consistent improvement.

\begin{figure*}[t]
    \centering
    \includegraphics[width=0.9\linewidth]{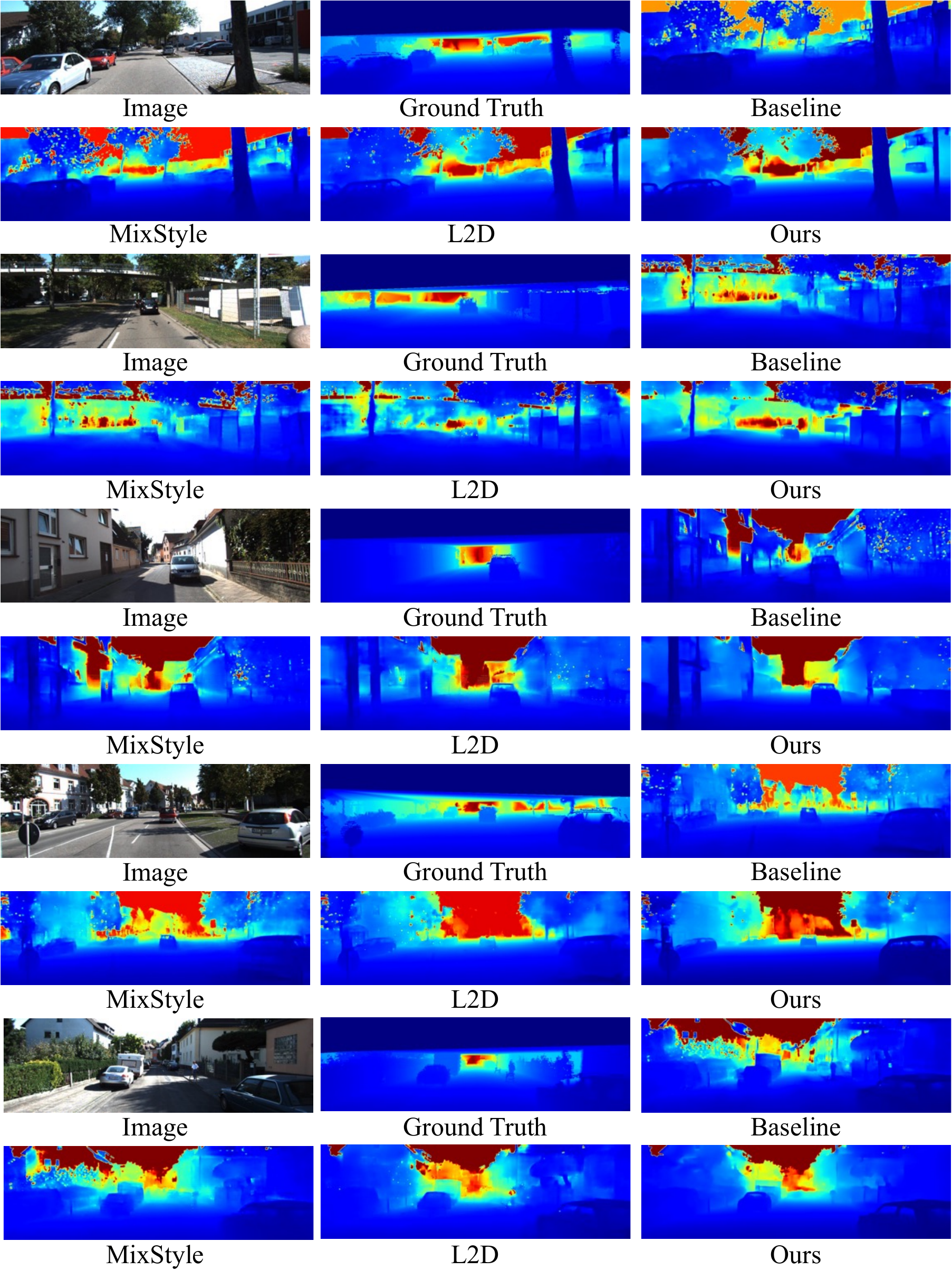}
    \caption{Qualitative comparisons on the monocular depth estimation task. The results of our method demonstrate clearer object boundaries with the least noise.}
    \label{fig:depth}
\end{figure*}

\noindent\textbf{Comparison on Monocular Depth Estimation.} 
As a supplement to the quantitative results of monocular depth estimation in the main manuscript, we provide qualitative comparisons with those methods without restriction on the label format, including MixStyle~\cite{zhou2021domain} and a variant of L2D~\cite{wang2021learning} by removing the class-conditional terms that are incompatible with this task, denoted as v-L2D.  
Using \textit{vKITTI}~\cite{gaidon2016virtual} as the source domain, results on the test KITTI dataset~\cite{geiger2013vision} are shown in  \cref{fig:depth}. 
Through the figure, we can observe that the results of our method show clear object boundaries with the least noise, which demonstrates the superiority of our method over the previous approaches. 

\begin{table}[t]
\scriptsize
\centering
    \begin{tabular}{ccccccc}
        \toprule
        Method & C & I & P & Q & S & Avg. \\
        \midrule
        L2D & 38.69 & 12.04 & 38.40 & 6.53 & 30.12 & 25.16 \\
        Ours & \textbf{42.93} & \textbf{12.79} & \textbf{40.72} & \textbf{6.78} & \textbf{32.67} & \textbf{27.18} \\
        \bottomrule
    \end{tabular}
    \caption{Comparisons of single DG accuracy (\%) on DomainNet. The models are trained on the Photo domain and evaluated on Clipart (C), Infograph (I), Painting (P), Quickdraw (Q), and Sketch (S). Results under the ResNet-18 backbone are reported.}
    \label{tab:domainnet}
\end{table}

\end{document}